\documentclass[journal]{IEEEtran}
\IEEEoverridecommandlockouts
\usepackage[utf8]{inputenc}
\usepackage[T1]{fontenc}
\usepackage{newunicodechar}
\newunicodechar{✓}{\checkmark}
\usepackage{cite}
\usepackage{amsmath,amssymb,amsfonts}
\usepackage{algorithmic}
\usepackage[ruled]{algorithm2e}
\usepackage{graphicx}
\usepackage{caption}
\usepackage{subcaption}
\usepackage{textcomp}
\usepackage{xcolor}
\usepackage{textcomp}
\usepackage{anyfontsize}
\usepackage{booktabs}
\usepackage{siunitx}
\usepackage[super]{nth}
\usepackage[para]{threeparttable}
\graphicspath{{./figs/}{.}}
\usepackage{svg}

\usepackage{ragged2e}
\usepackage{dirtree}

\usepackage[acronym, style=super, nonumberlist]{glossaries}
\newacronym{adc}{ADC}{analog-to-digital converter}
\newacronym{cnn}{CNN}{convolutional neural network}
\newacronym{cpc}{CPC}{conductive polymer composite}
\newacronym{dma}{DMA}{direct memory access}
\newacronym{eth}{ETH}{Eidgenössische Technische Hochschule}
\newacronym{fsr}{FSR}{force sensitive resistor}
\newacronym{fps}{FPS}{frames per second}
\newacronym{gpu}{GPU}{graphics processing unit}
\newacronym{gui}{GUI}{graphical user interface}
\newacronym{i2c}{I2C}{inter integrated circuit}
\newacronym{ic}{IC}{integrated circuit}
\newacronym{iis}{IIS}{Integrated Systems Laboratory}
\newacronym{imu}{IMU}{inertial measurement unit}
\newacronym{led}{LED}{light-emitting diode}
\newacronym{ldo}{LDO}{low-dropout}
\newacronym{macc}{MACC}{multiply-and-accumulate operation}
\newacronym{mcu}{MCU}{microcontroller unit}
\newacronym{mems}{MEMS}{microelectromechanical system}
\newacronym{mit}{MIT}{Massachusetts Institute of Technology}
\newacronym{mlp}{MLP}{multi-layer perceptron}
\newacronym{pcb}{PCB}{printed circuit board}
\newacronym{pulp}{PULP}{parallel ultra low power}
\newacronym{ram}{RAM}{random-access memory}
\newacronym{sdram}{SDRAM}{synchronous dynamic random-access memory}
\newacronym{spdt}{SPDT}{single pole double throw}
\newacronym{spi}{SPI}{serial peripheral interface}
\newacronym{stag}{STAG}{scalable tactile glove}
\newacronym{tcn}{TCN}{temporal convolutional neural network}
\newacronym{uart}{UART}{universal asynchronous receiver-transmitter}
\newacronym{wom}{WOM}{wake-on-motion}
\newacronym{afe}{AFE}{analog front-end}
\newacronym{cv}{CV}{cross-validation}

\usepackage{tikz}
\usepackage{pgfplots}
\pgfplotsset{width=7cm,compat=1.8}
\usepgfplotslibrary{fillbetween}

\usepackage{url}

\newcommand{\new}[1]{{\color{black}#1}}
\newcommand{\rebuttal}[1]{{\color{black}#1}}
\newcommand{\vlad}[1]{{\color{black}#1}}

\usepackage{fancyhdr}
\fancypagestyle{mahmood}{%
  \fancyhf{} 
  
  \fancyhead[R]{1570720892}
}%
\makeatletter
\let\ps@IEEEtitlepagestyle\ps@mahmood
\makeatother

\usepackage{fancyhdr}
\fancypagestyle{mahmood}{%
  \fancyhf{} 
  
  \fancyhead[C]{\footnotesize \textcopyright 2022 IEEE. Personal use of this material is permitted.  Permission from IEEE must be obtained for all other uses, in any current or future media, including reprinting/republishing this material for advertising or promotional purposes, creating new collective works, for resale or redistribution to servers or lists, or reuse of any copyrighted component of this work in other works.}
}%
\makeatletter
\let\ps@IEEEtitlepagestyle\ps@mahmood
\makeatother

\begin{document}


\title{
\fontsize{23}{28}\selectfont
Leveraging Tactile Sensors for Low Latency Embedded Smart Hands for Prosthetic and Robotic Applications
}
\author{Xiaying~Wang,~\IEEEmembership{Student Member,~IEEE,}
        Fabian~Geiger,
        Vlad Niculescu,
        Michele~Magno,~\IEEEmembership{Senior Member,~IEEE,}
        Luca~Benini,~\IEEEmembership{Fellow,~IEEE}
\thanks{Manuscript received November 07, 2019; revised January 16, 2020; accepted February 10, 2020.}
\thanks{All authors are affiliated with the Department of Information Technology and Electrical Engineering, ETH Z{\"u}rich, Switzerland (Corresponding e-mail: xiaywang@iis.ee.ethz.ch).}
\thanks{This project was supported by the Swiss Data Science Center PhD Fellowship under grant ID P18-04.}
    }
\maketitle

\begin{abstract}

Tactile sensing is a crucial perception mode for robots and human amputees in need of controlling a prosthetic device. Today robotic and prosthetic systems are still missing the important feature of accurate tactile sensing. This lack is mainly due to the fact that the existing tactile technologies have limited spatial and temporal resolution and are either expensive or not scalable.
In this paper, we present the design and the implementation of a hardware-software embedded system called SmartHand. It is specifically designed to enable the acquisition and the real-time processing of high-resolution tactile information from a hand-shaped multi-sensor array for prosthetic and robotic applications. 
During data collection, our system can deliver a high throughput of 100 frames per second, which is 13.7$\times$ higher than previous related work. This has allowed the collection of a new tactile dataset consisting of 340,000 frames while interacting with 16 objects from everyday life during five different sessions. Together with the empty hand, the dataset presents a total of 17 classes. 
We propose a compact yet accurate convolutional neural network that requires one order of magnitude less memory and 15.6$\times$ fewer computations compared to related work without degrading classification accuracy.
The top-1 and top-3 cross-validation accuracies on the collected dataset are respectively 98.86\% and 99.83\%. We further analyze the inter-session variability and obtain the \rebuttal{best top-3} leave-one-out-validation accuracy of 77.84\%.
We deploy the trained model on a high-performance ARM Cortex-M7 microcontroller achieving an inference time of only 100\,ms minimizing the response latency. The overall measured power consumption is 505\,mW. Finally, we fabricate a new control sensor and perform additional experiments to provide analyses on sensor degradation and slip detection.
This work is a step forward in giving robotic and prosthetic devices a sense of touch by demonstrating the practicality of a smart embedded system that utilizes a scalable tactile sensor with embedded tiny machine learning. 

\end{abstract}


\section{Introduction}
\label{ch:introduction}

Recent advances in electronics and \gls{mems} manufacturing have allowed the sophistication of sensors and robotic devices to grow by leaps and bounds \cite{jiang2021tactile,saadatzi2019modeling,song2021bionic}.
\rebuttal{Robotic hands and prosthetic devices provide robots with humans' ability to interact with the surroundings and restore lost abilities for amputees~\cite{Meattini2018sEMG,precup2020evolving, Santoni2019roboticarmtim,song2021bionic}.}
The preeminent modality of choice for robotic devices that need to interact with the environment is vision~\cite{BERGAMINI2020101052,Cheng2020visiontim,sahoo2019geometry}. 
Optical sensors are performing well, are inexpensive, and mature computer vision algorithms exist for every level of processing power.
On the other hand, there are tasks in which optical information is insufficient or leads to disproportionately complex control strategies.
One such task is the manipulation of arbitrary objects with an articulated end-effector. Vision is required to find correctly the object and the position of the end-effector, but once contact is made, the robot cannot tell if the object is fragile or robust, if the end-effector has a good grip or if the object will slip from its grasp. For this reason, a recent trend is to develop non-visual technologies and methods to provide a robot with those capabilities. 

In nature, organisms have evolved a useful modality for manipulating their surroundings: the touch.
Science is often inspired by nature, so in order to improve the ability of robots and prosthetic devices to manipulate objects in the environment, tactile sensors have recently gained considerable attention.
Even though much research has been conducted~\cite{weiner2020embeddedfingers, Zou2017, Kappassov2015, Xu2018, maddipatla2017pressuresensor}, the task of developing a tactile sensor with a resolution comparable to the human hand, with similar robustness as well as flexibility, remains an open challenge~\cite{Sundaram2019}.
Object grasping and manipulation are natural abilities for healthy human beings, but it is a very difficult task for robotic machines. Besides the sense of touch, humans can elaborate sensory and perceptive inputs during object interaction. The inertial properties of the object are immediately perceived, and the corresponding adjustments are performed in real-time during the manipulation.
%
This is an essential ability that a robotic machine has to be equipped with in order to optimally interact with objects. In fact, the weight, the mass distribution, and the inertial resistance are all factors to be considered for efficient and safe robotic motor planning. 
However, no satisfying solution has yet been proposed due to the lack of well-defined sensing equipment that provides useful data for estimating the inertial parameters~\cite{MAVRAKIS2020inertial}. 

State-of-the-art sensor solutions, from academia and industry, are typically hard to manufacture, expensive, and lacking in temporal as well as spatial resolution~\cite{Yin2018, Zhang2018, Franceschi2017}. The majority of the solutions mainly focus on replicating the fingers tips~\cite{syntouch,weiner2020embeddedfingers,Choi2020finger,song2021bionic} or have sensors covering only a limited part of the hand palm~\cite{Kang2017touchsensor,song2021bionic,jiang2021tactile,saadatzi2019modeling,Altamirano2020palm, abbass2021novel}. 
\vlad{While \cite{zhu2020haptic} presents a system with full hand coverage, they use a small number of sensing elements (i.e., 16) and therefore only prove the functionality of the system with seven objects.
}
Due to the high cost of the commercial solutions and the small size of the most common solution usually covering only a single finger, the access to tactile information for small research groups or companies is still very limited and it is not unleashing the potential of the research community~\cite{narang2021interpreting}.
\rebuttal{In contrast, Sundaram and colleagues have proposed a full-hand sensor using an affordable technology~\cite{Sundaram2019}.} The authors have designed and developed a \gls{stag} enabling the full coverage of the human hand, providing tactile information with high spatial resolution.
The preliminary work of the tactile glove includes a dataset collected while interacting with 26 daily life objects for several minutes each. \rebuttal{A \gls{cnn} is subsequently applied for classifying the objects and the empty hand, i.e., 27 classes.
The top-1 and the top-3 accuracy results are around 74\% and 89\%, respectively.} 
To be noted that the proposed setup with offline sensor data processing based on several-minute recordings introduces long latency, making it sub-optimal for a real-world application where a prompt response in the range of milliseconds is desired. 
Moreover, embedding multi-sensors in small integrated systems is actively researched thanks to its advantages for closed-loop control of the robotic devices~\cite{Yunas2020sas}. Weiner and colleagues~\cite{weiner2020embeddedfingers} propose a fingertip system integrating several sensors that are miniaturized to fit in the very limited space of a fingertip. Besides the integration of sensors, a smart processing unit, e.g., \glspl{mcu}, that can process the acquired sensor data at the edge would enable a real-time response of the whole system.
\vlad{Table~\ref{tab:related_works} provides an in-depth comparison of the mentioned related works in terms of hand coverage, number of sensors, detection latency and classification performance.}

\rebuttal{Machine learning and deep learning have been widely applied in medical field.}
\vlad{For instance, the authors in \cite{roy2018improving} propose a method for detecting and eliminating the artifacts in the photoplethysmographic measurements using artificial neural networks (ANNs).
Furthermore,  \cite{albu2019results} also takes advantage of ANNs and presents a system that could assist the physician in making a decision on the current state of the patient or in predicting the future states.}
Recent developments in edge processing units and tinyML~\cite{wang2020fann} make it possible to bring the data processing close to the sensors~\cite{zhao2021weighted}. 
\vlad{Along with efforts in designing and deploying tiny machine learning and deep learning models on the edge, nowadays edge devices are becoming increasingly smart and can interpret sensor data and translate it into actionable information in real-time, for example gesture recognition using radar~\cite{scherer2021radar,kang2021measurement} and brain--machine interfaces using electroencephalograms~\cite{wang2020accurate}.}
A drawback of low-power edge computing is that the resources available on \glspl{mcu} with milli-watt power consumption are very limited making it impossible to deploy big and complex models. Thus, a conscious design of tiny yet accurate models is necessary, by taking into consideration the memory availability and the computational capability of the underlying processor.

In this work, we propose a step towards equipping robotic and prosthetic devices with a smart sense of touch by presenting a smart embedded system, called Smarthand, capable of reading and processing high-resolution tactile information in real-time. A preliminary version of the Smarthand has been presented in~\cite{wang2021smarthand}. 
This paper significantly extend the previous work including a more extensive methodology and new technical contributions. We present for the first time a novel read-out front-end to improve the acquisition of the data in terms of energy and datarate. 
The experimental results present more accurate and extensive evaluations on latency, power, and accuracy of the proposed solution. 
Furthermore, we fabricate a new control sensor and provide quantitative and qualitative analyses on sensor degradation and slip evaluations.

The main contributions of this paper are as follows:
\begin{itemize}
    \item We extend the SmartHand~\cite{wang2021smarthand} with a novel front-end able to acquire tactile and hand movement data with a high throughput of 100 frames per second, which is 13.7$\times$ higher than~\cite{Sundaram2019}, and a power consumption of 52\,mW.
    \item We present a more extensive evaluation of the novel \gls{cnn} algorithm to classify 16 objects and the empty hand considering the inter-session variability. We deploy the model on a ARM Cortex-M7 \gls{mcu} reaching a low inference latency of 100\,ms and a power consumption of 430\,mW.
    \item We fabricate a new square sensor and perform controlled experiments to assess quantitatively and qualitatively the sensor degradation between sessions and slip behaviours. Experimental results show that the average response of all contact-frames diminishes by up to 40\% for the sensor glove and almost 30\% for the control sensor.
    \item Experimental power measurements demonstrate that our proposed solution consumes 505\,mW in active mode and 185\,\textmu W in standby mode. Assuming a duty cycle of 10\% and a battery capacity of 1\,Wh, our proposed SmartHand can deliver an operating time of up to 20 hours per day.
\end{itemize}
\rebuttal{We open-source release the dataset and the codes\footnote{\label{smarthand_url}\url{https://iis.ee.ethz.ch/~datasets/smarthand/}}.}

\begin{table}[]
\setlength{\tabcolsep}{4.2pt}
\caption{\vlad{Comparison to related works. N/A: Not Applicable. N/S: Not Specified.}}
\label{tab:related_works}
\begin{tabular}{@{}lrrrrrrr@{}}
\toprule
\multicolumn{1}{l}{}                                               & \multicolumn{1}{r}{\cite{jiang2021tactile}} & \multicolumn{1}{r}{\cite{abbass2021novel}}                                          & \multicolumn{1}{r}{\cite{Sundaram2019}}                                 & \multicolumn{1}{r}{\cite{Kang2017touchsensor}} & \multicolumn{1}{r}{\cite{Altamirano2020palm}} & \multicolumn{1}{r}{\cite{zhu2020haptic}}                                 & \multicolumn{1}{r}{Ours} \\ \midrule
\begin{tabular}[c]{@{}l@{}}Hand \\ coverage\end{tabular}           & finger                & \begin{tabular}[c]{@{}r@{}}finger tips\\ and palm\end{tabular} & \begin{tabular}[c]{@{}r@{}}whole \\ hand\end{tabular} & palm                  & palm                  & \begin{tabular}[c]{@{}r@{}}whole \\ hand\end{tabular} & \begin{tabular}[c]{@{}r@{}}whole \\ hand\end{tabular} \\
\begin{tabular}[c]{@{}l@{}}\# of sensing \\ elements\end{tabular}  & 96                    & 64                                                             & 548                                                   & 64                    & 15                    & 16                                                    & 548                                                   \\
\begin{tabular}[c]{@{}l@{}}Latency\\ {[}ms{]}\end{tabular}         & 150                   & N/A                                                            & N/S                                                   & N/A                   & N/A                   & 40                                                    & 100                                                   \\
\begin{tabular}[c]{@{}l@{}}Classification\\ algorithm\end{tabular} & yes                   & no                                                             & yes                                                   & no                    & yes                   & yes                                                   & yes                                                   \\
\# of classes                                                      & N/S                   & N/A                                                            & 27                                                    & N/A                   & 4                     & 7                                                     & 17                                                    \\
\begin{tabular}[c]{@{}l@{}}Classification\\ accuracy\end{tabular}  & N/S                   & N/A                                                            & 90\%                                                  & N/A                   & 84\%                  & 96\%                                                  & 99\%                                                  \\
\begin{tabular}[c]{@{}l@{}}On-board\\ processing\end{tabular}      & yes                   & N/A                                                            & no                                                    & N/A                   & no                    & N/S                                                   & yes                                                   \\
\begin{tabular}[c]{@{}l@{}}Sampling\\ resolution\end{tabular}      & 12-bit                & N/S                                                            & 10-bit                                                & N/S                   & N/S                   & N/S                                                   & 12-bit                                                \\ \bottomrule
\end{tabular}
\end{table}
\section{Background}
\label{sec:bg}

\rebuttal{This section provides related background knowledge for understanding this paper.}

\new{

\subsection{Tactile Sensing}\label{bg:tactilesensing}
Tactile sensing requires that different tactile cues are transduced to an electrically detectable signal.
The force sensitivity of a \gls{cpc}~\cite{CPCphysics} acts as a \gls{fsr} when it is sandwiched between two electrodes.
The resistance of such a sensor sandwich consists of two parts: a contact resistance and a polymer resistance.
The contact resistance depends on the type of electrodes used, their shape and how they are bonded to the polymer.
It is ideally constant for a certain setup and does not drift or change over time.
Whereas, the polymer resistance transduces variations in applied force to variations in resistance.
\rebuttal{The resistance value and its behavior depend on the concentration of conductive particles that is added to the fabrication of the \gls{cpc}.
The resistance decreases with incremental applied stress if the concentration is below the percolation threshold, while the opposite is true above the threshold.}
An overview over the physical properties of conductive polymer composites can be found in~\cite{CPCphysics}.

\subsection{Related Dataset}

\rebuttal{Sundaram and colleagues~\cite{Sundaram2019} have proposed for the first time a tactile glove with high spatial resolution covering the full hand, called \gls{stag}, based on the concept of \gls{fsr} using a \gls{cpc}.} 
A total of 135,000 frames of tactile data has been collected together with synchronized video footage in three recording sessions, during each of which the sensor glove is worn throughout the entire session.
This dataset contains several minutes of interaction with 26 different objects from everyday life. 
Tactile frames are collected at a rate of about 7.3\,Hz. All objects are manipulated in a repeatable way, mostly pushing the sensor from the top onto the object lying on a table.
The frames corresponding to the actual touch of the object are chosen by comparing the values with a threshold found from an empty-hand dataset.
This threshold value is found for each taxel to account for any internal stress caused by the sensor conformation. 
The authors have selected N frames from the several-minute recordings by either random selection or clustering to construct the input samples to a neural network for object classification.
The dataset is publicly available\footnote{\url{http://stag.csail.mit.edu/}} \rebuttal{and we name it MIT-STAG dataset in this paper for clarity.}

}

\section{Methods}\label{sec:methods}

\rebuttal{This section explains the proposed system architecture and its operational modalities in Sec.~\ref{ch:sysarch} and Sec.~\ref{subsec:data_collection}, respectively, the recorded dataset in Sec.~\ref{subsec:our_dataset}, the proposed neural network, the validation methodologies, and its embedded implementation in Sec.~\ref{sec:cnn}. Finally, the additional evaluations of the tactile sensor are explained in Sec.~\ref{subsec:methods:sensoreval}. The experimental results are reported and discussed in Sec.~\ref{ch:results}.}

\subsection{System Architecture} \label{ch:sysarch}

\begin{figure}[t]
  \centering
  \includegraphics[width=.95\linewidth]{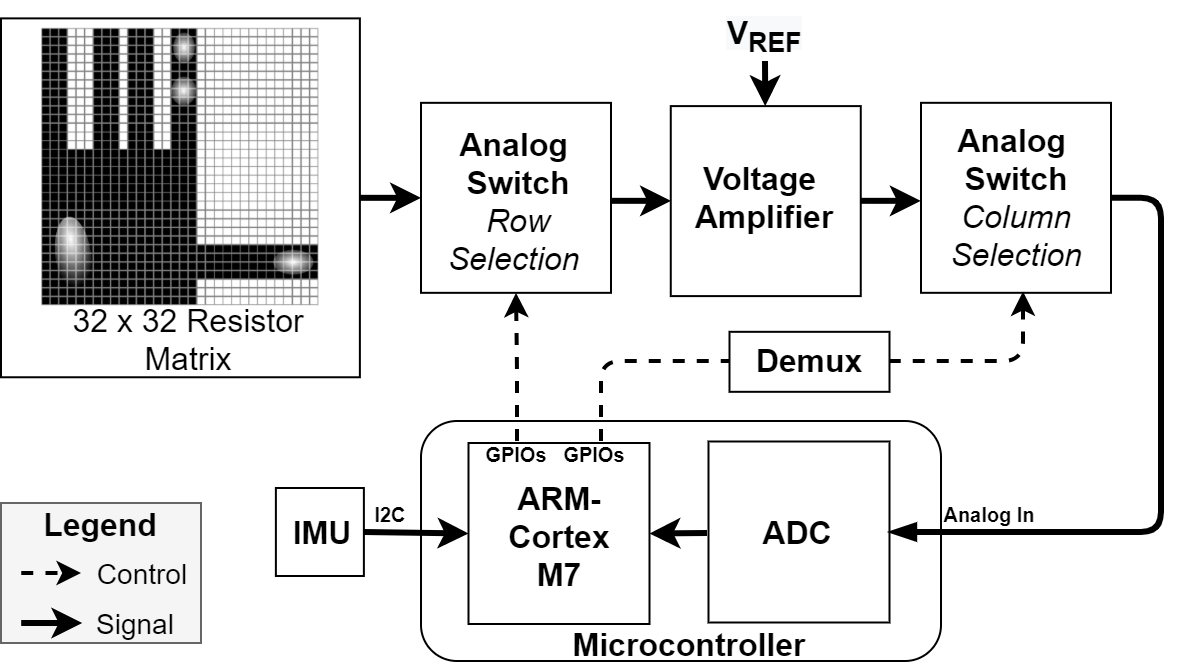}
  \caption{The block diagram of the system.}%
  \label{fig:block}
\end{figure}

\begin{figure}[t]
  \centering
  \includegraphics[trim={0cm 10cm 13cm 6cm},clip, width=.75\linewidth]{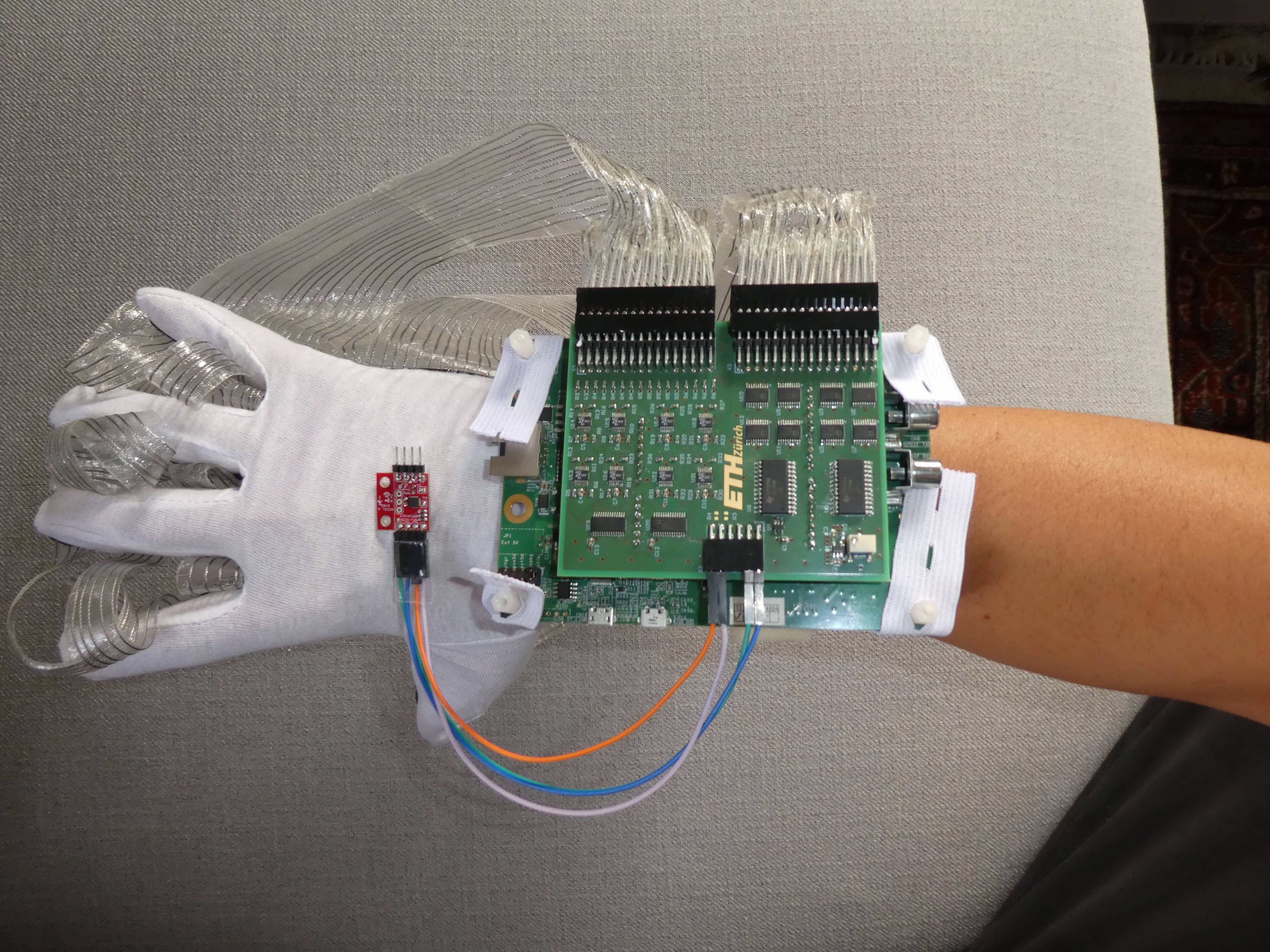}
  \caption{Fully assembled system from the top. The red PCB on the back of the glove is the IMU. The readout circuit is plugged onto the discovery board like a shield.}%
  \label{fig:full_system_top}
\end{figure}

\begin{figure}[t]
  \centering
  \includegraphics[trim={5cm 10cm 5cm 0},clip, width=.75\linewidth]{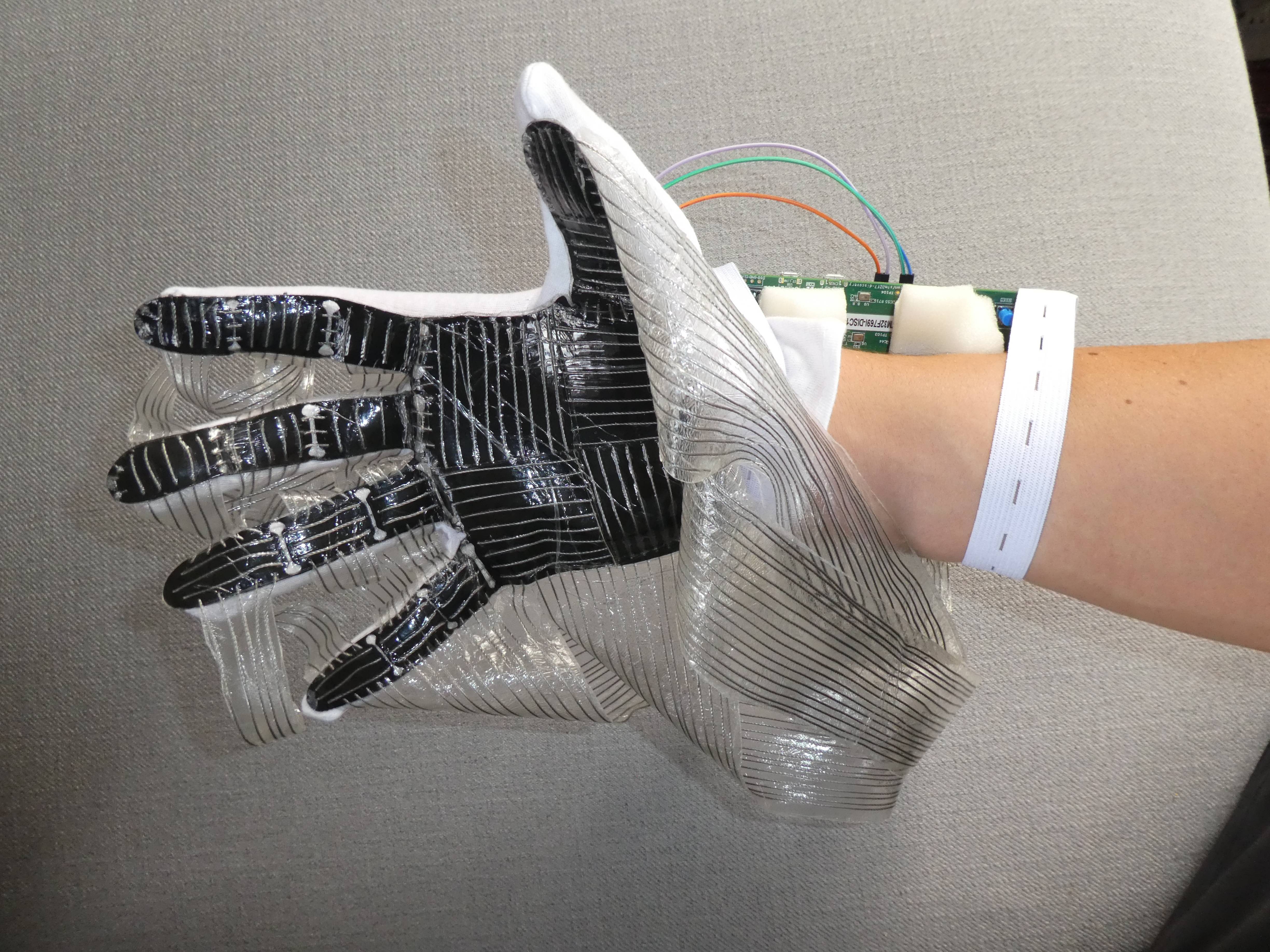}
  \caption{Fully assembled system from the bottom. The sensor laminate is glued to the white cotton glove.}%
  \label{fig:full_system_bot}
\end{figure}

Fig.~\ref{fig:block} depicts the block diagram and the architecture of the designed Smarthand. It consists of a low cost resistive  tactile sensor based on a \gls{cpc} as in~\cite{Sundaram2019}, an \gls{afe} to read the tactile data, an \gls{imu} to gain additional movement information, and an \gls{mcu} for system control and on-board signal processing.
Fig.~\ref{fig:full_system_top} and Fig.~\ref{fig:full_system_bot} show the full system assembled and worn.


\subsubsection{Tactile sensor}
\begin{figure}[t]
  \centering
  \includegraphics[width=.65\linewidth]{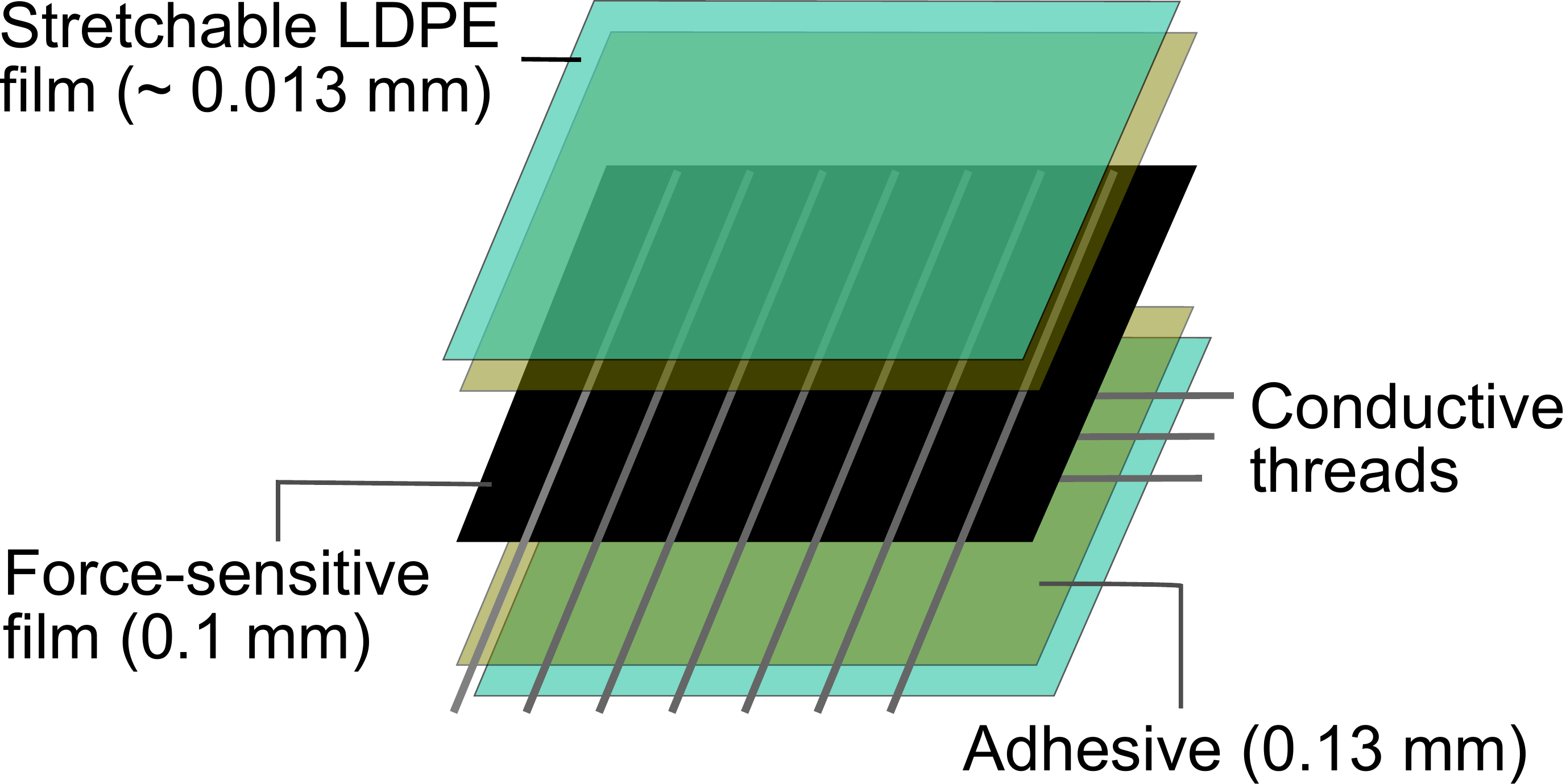}
  \caption{Sensor composition. Adhesive and low-density polyethylene (LDPE) film are required to bond the electrodes to the force sensitive film and protect it, respectively~\cite{Sundaram2019}.}%
  \label{fig:sensor_structure}
\end{figure}

Following the fabrication procedures described in~\cite{Sundaram2019}, a tactile sensor with high spatial resolution has been produced in our laboratory. 
\new{The sensor is based on the concept of \gls{fsr}, as explained in Sec.~\ref{bg:tactilesensing}. It transduces changes in applied force or pressure to changes in resistance that can be read with a dedicated analog front-end.}
It consists of a \gls{cpc}, called Velostat, sandwiched between two orthogonal sets of electrodes, as shown in Fig.~\ref{fig:sensor_structure}.
\new{The conductive particles in Velostat are carbon black and their concentration is below the percolation threshold, so its resistance decreases in response to increasing normal forces.
The sensor electrodes are smooth stainless steel threads with a thickness of 0.25mm and there are 32 electrodes running across each of the two sides of the force sensitive film forming a grid.}

\begin{figure}[t]
  \centering
  \includegraphics[width=.75\linewidth]{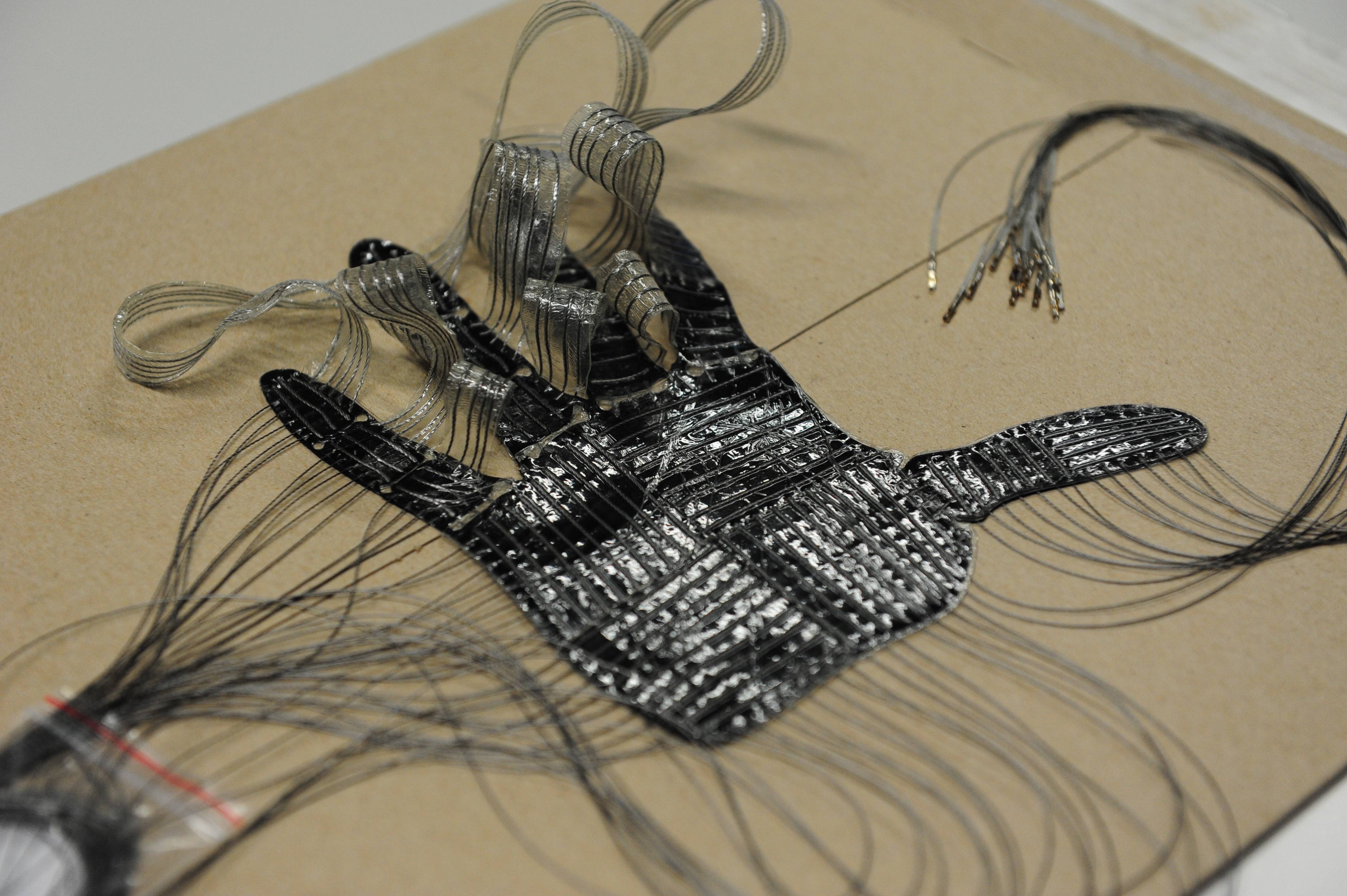}
  \caption{Fabricated sensor before being glued to a glove.}%
  \label{fig:sensor_thesis}
\end{figure}

\begin{figure}[t]
  \centering
  \includegraphics[width=.75\linewidth]{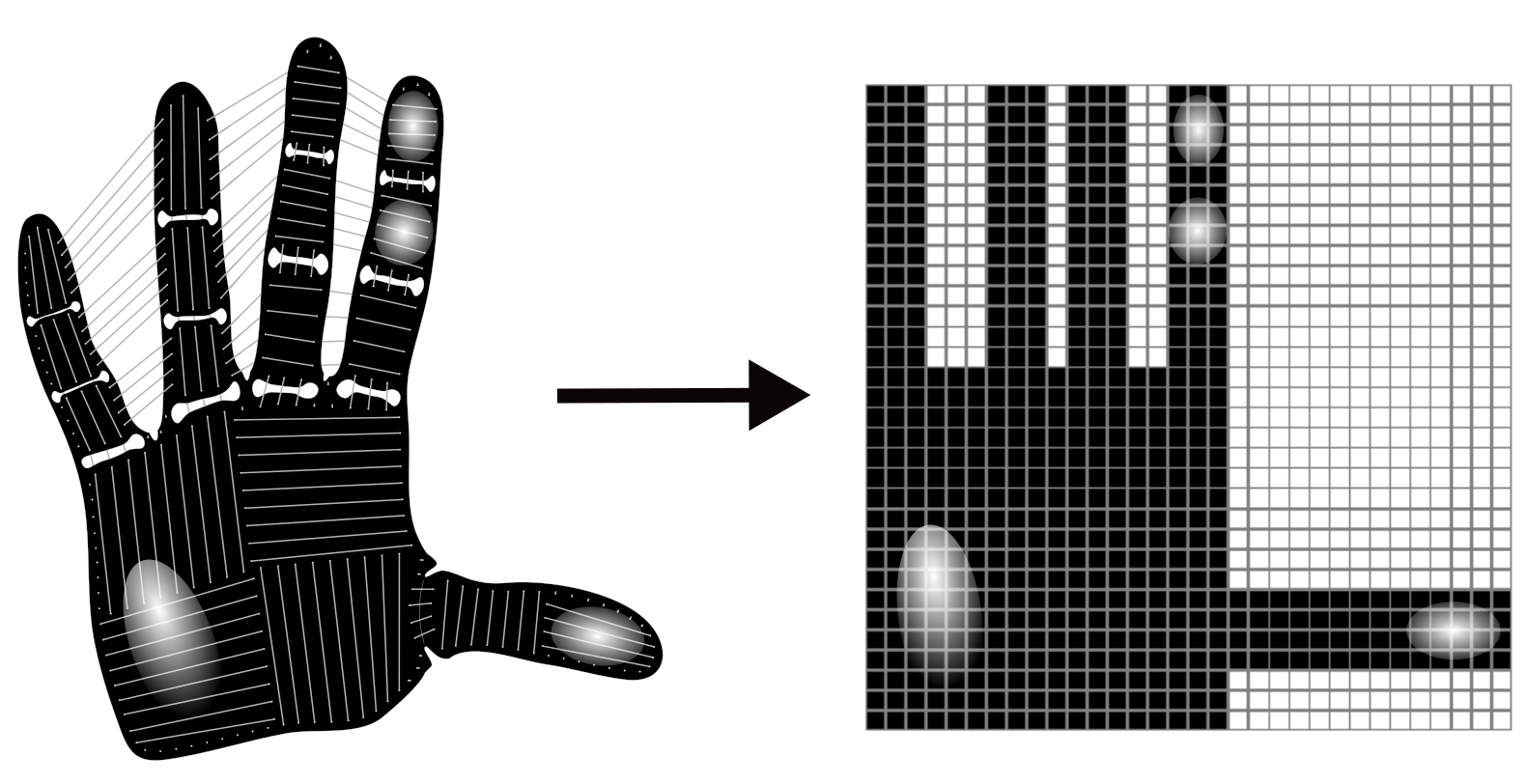}
  \caption{Representation of the sensor values as a 32x32 matrix, referred to as \emph{frame} or \emph{tactile frame}.}%
  \label{fig:glove2frame}
\end{figure}

The function of the sensor does not depend on its shape, thus it can be fabricated in arbitrary forms and sizes.
To obtain comparable results and capture useful data during object interaction, the sensor has been fabricated as similarly as possible to~\cite{Sundaram2019}. Fig.~\ref{fig:sensor_thesis} shows the finished sensor laminate before being attached to a thin cotton glove.
Because the active sensor area is shaped like a hand, there are 548 physical crossings of electrodes.
However, to preserve the spatial relation between taxels, or sensor points, and make it easier to work with the sensor data, all 1024 available taxels are read.
By arranging the read values in a 32$\times$32 matrix, an intuitive and natural representation of the sensor can be constructed, as shown in Fig.~\ref{fig:glove2frame}, and will be referred to as \emph{tactile frame}, or simply \emph{frame}.

\subsubsection{Readout circuit}
\label{subsec:readout_circuit}

\begin{figure}[t]
  \centering
  \includegraphics[width=.95\linewidth]{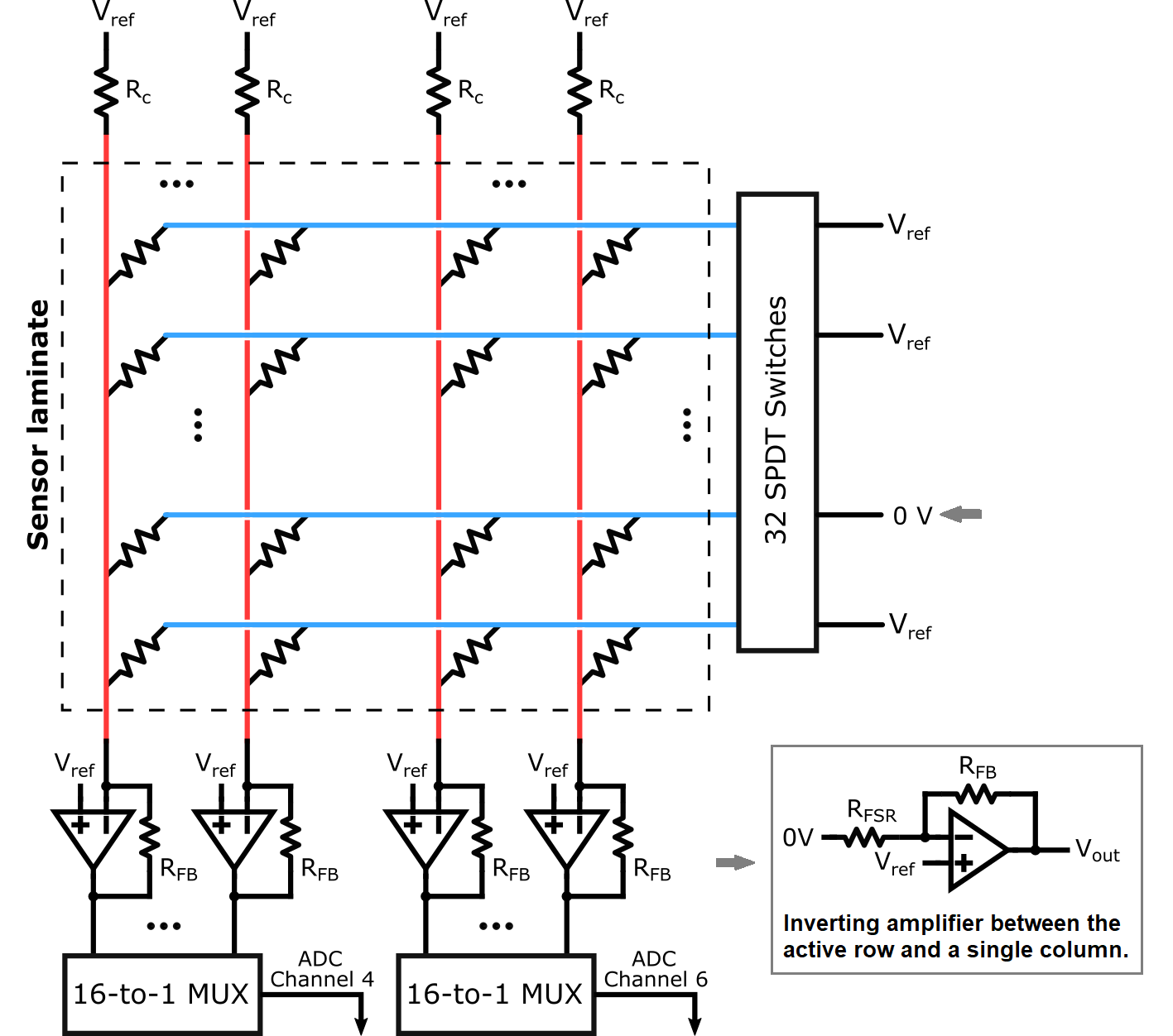}
  \caption{\new{Block diagram of the sensor with its readout circuit.}}%
  \label{fig:readout_circuit}
\end{figure}

\new{
Abstracting from its physical structure, the tactile sensor can be drawn as a simple resistor network with variable resistances, formed by the cross-talks between the two orthogonal sets of 32 electrodes. 
Such an abstract representation is shown in Fig.~\ref{fig:readout_circuit} framed with dashed line.}
Each cross-talk can be seen as a simple resistor with variable resistances depending on the pressure applied to the sensor glove.
\new{The biggest challenge in accurately measuring each resistor in this matrix configuration is posed by the extensive cross-talk between electrodes.

There are two main approaches for handling cross-talk.
The first one makes sure that there is only one current path when measuring the resistance between two electrodes, which is through that same resistor~\cite{Kim2016}.
Thus, it can also be described as electrical isolation scheme.
The second approach takes multiple measurements, including cross-talk, and finds the true resistances by solving a set of linear equations~\cite{Hindawi2018}.
The advantage of this approach is that a minimum number of components is required to realize the readout circuit.
However, it is an iterative procedure so it is computationally more expensive and slower than the first approach.
Since any \gls{mcu} should be able to efficiently use the proposed system, we have decided to implement a variation of the electrical isolation scheme for a fast readout.

{We base our implementation on the readout circuit used in~\cite{Sundaram2019} and optimize it for low-power, speed, and accuracy by using a more powerful \gls{mcu} featuring a higher precision 12-bit \gls{adc}.} 
The working principle is that one row is grounded, while all other rows and columns are pulled up to a reference voltage.
This ideally prevents cross-talk between rows, and gives the opportunity to read out all columns of the active row at the same time.
The active row and all columns form an inverting amplifier configuration, with the \gls{fsr} of the Velostat in the forward path and a simple resistor in the feedback path.
The positive input of the operational amplifier is directly connected to the reference voltage, resulting in the following equation for the output voltage:
\begin{equation}
  V\sb{out} = V\sb{out} \cdot \frac{R\sb{FB}+R\sb{FSR}}{R\sb{FSR}}
\end{equation}

\rebuttal{A block diagram of the sensor, its readout circuit, and the inverting amplifier configuration between the active row and a single column is shown in Fig.~\ref{fig:readout_circuit}.}
}

\subsubsection{Movement sensor}
We additionally integrate an \gls{imu} for acquiring accelerometer and gyroscope data to collect hand movements during object interaction.
The used \gls{imu} is \mbox{MPU-9250} with a sampling frequency of \SI{100}{\hertz} and a resolution of 16-bit for both the accelerometer and the gyroscope.

\subsubsection{Edge processing unit}
The embedded system requires an \gls{mcu} with the ability to drive and read the tactile sensor, as well as directly run neural network inferences with the acquired data and take action based on the inference, e.g., actuating the motors of the robotic arm.
For a fast development of the prototype, we have chosen the STM32F769NI discovery board, featuring an ARM Cortex-M7 core. With its 2\,MB of Flash memory, 532\,kB of RAM, and a maximum clock speed of 216\,MHz, it is one of the most highly-performing \gls{mcu} in the low-power ARM Cortex-M family. 
It comes with the software tool STM32CubeMX, from STMicroelectronics, employed for generating initialization code, compiling, and flashing.
\mbox{X-CUBE-AI} is an expansion package for STM32CubeMx that allows fast deployment of neural networks on STM32 \glspl{mcu}.

\subsection{Operational Modalities}\label{subsec:data_collection}
The firmware of the system is implemented such that three operational modalities are available for different use-case scenarios. A basic timer peripheral on the \gls{mcu} is used to clock the applications.

\paragraph{Data collection}
This modality is used for acquiring a complete dataset. The system waits for a `start' command, given either by pressing a button on the discovery board or sending the character `r' via the \gls{uart} protocol to the \gls{mcu}. 
If either of these two events is detected, the base timer is triggered at a rate of 100\,Hz, and the tactile along with \gls{imu} data is collected.
The external \gls{sdram} on the discovery board was utilized as intermediate storage for tactile data, while the \gls{imu} data is sent directly after its acquisition. 
After a configurable number of frames has been collected, the data acquisition stops and immediately sends the content of the \gls{sdram} to a connected computer, and the system returns to the idle state waiting for the next `start' command.
The \acrshort{sdram} can store a maximum of 4096 tactile frames, limiting one single interaction to a maximum of about 40 seconds with a data collection rate of 100Hz. The acquisition frequency can be reduced if a longer recording is desired.

\paragraph{Real-time data visualization}
A \gls{gui} designed using Python is implemented and can be used to visualize the tactile sensor data on a computer screen directly. Once the `start' command is received, the base timer is activated, and the tactile frames are collected at 10\,Hz and directly sent to the connected computer via \acrshort{uart}. To stop the visualization, either the button on the discovery board has to be pressed again, or the character `p' needs to be sent to the system.
The \gls{gui} displays each frame as a 32x32 matrix and provides two buttons, labeled `Run' and `Pause', which respectively send `r' or `p' to the system, giving the option to control the data visualization from the same interface.

\begin{figure}[t]
  \centering
  \includegraphics[trim={11cm 0 14cm 0},clip, width=.75\linewidth]{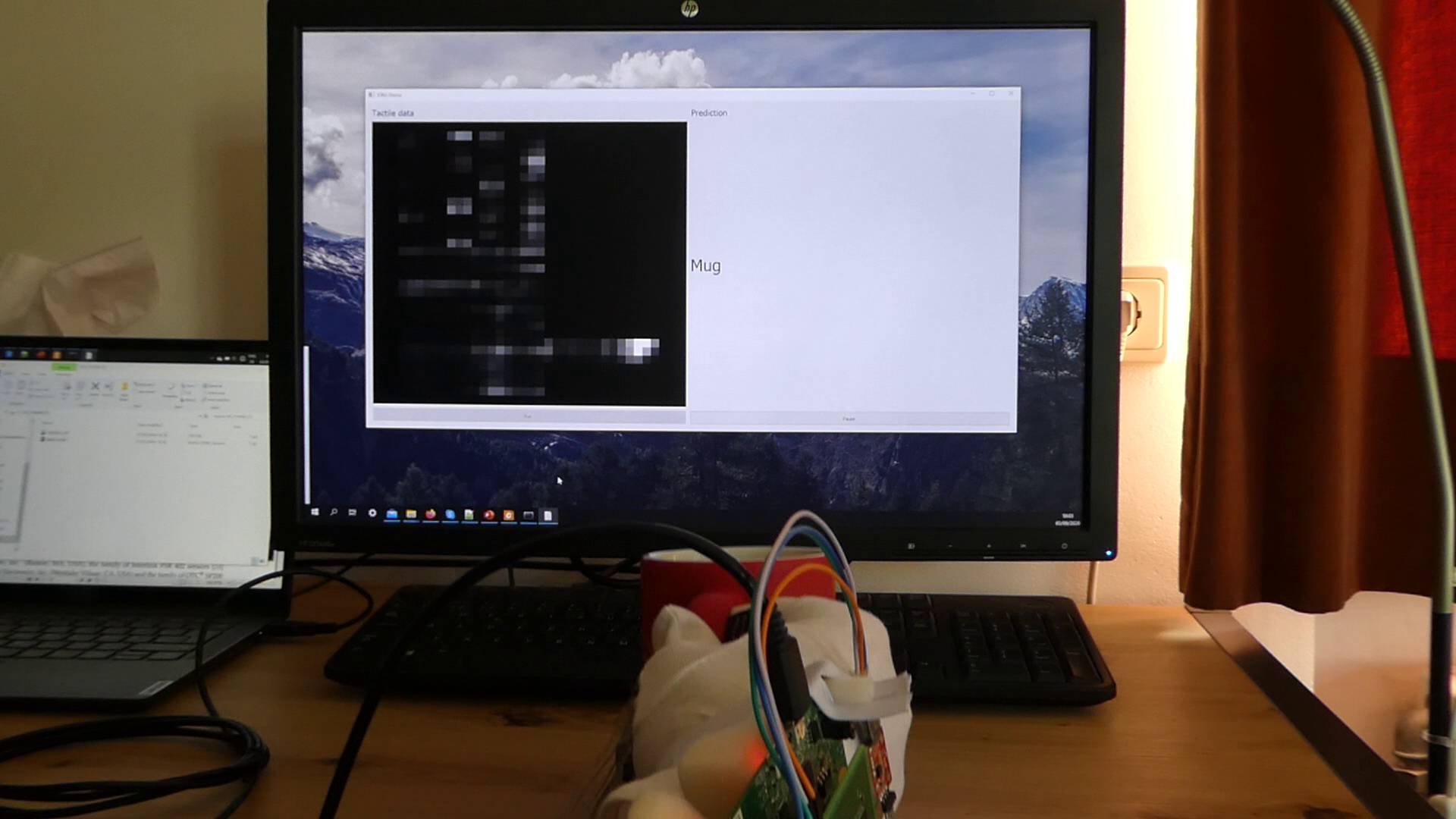}
  \caption{Snapshot from a demo video in full-system modality.}%
  \label{fig:demo_snapshot}
\end{figure}

\paragraph{Smarthand system}
This modality leverages the full system by not only reading tactile data but also processing it real-time on the \gls{mcu} using a deployed neural network. 
The data acquisition starts after pressing the button on the discovery board with the base timer configured to 8\,Hz. The sensor data is read and processed by the neural network on the \gls{mcu}. The system clock is configured to 216MHz to maximize the inference speed.
To evaluate the performance of the SmartHand system, a second Python \gls{gui} is developed, which displays the result of each network inference and its corresponding input frame. To simultaneously process the frame on-board and visualize it in real-time, a \gls{dma} controller is used for sensor data transmission. It is to note that in this case, the connection to the computer serves merely for the verification purpose of a demo, the system can fully function independent of any processing engine other than the on-board \gls{mcu}, and the output of the neural network inference can be used in real-time for any robotic or prosthetic control. 
Fig.~\ref{fig:demo_snapshot} displays a snapshot of a demo video in this application modality. 
The SmartHand is worn like a glove, and the tactile data during the interaction with a mug is acquired and classified immediately with the on-board embedded \gls{cnn}. The classification output is displayed in real-time on the screen.

\subsection{ETHZ-STAG Dataset}\label{subsec:our_dataset}

\begin{figure}[t]
  \centering
  \includegraphics[width=.8\linewidth]{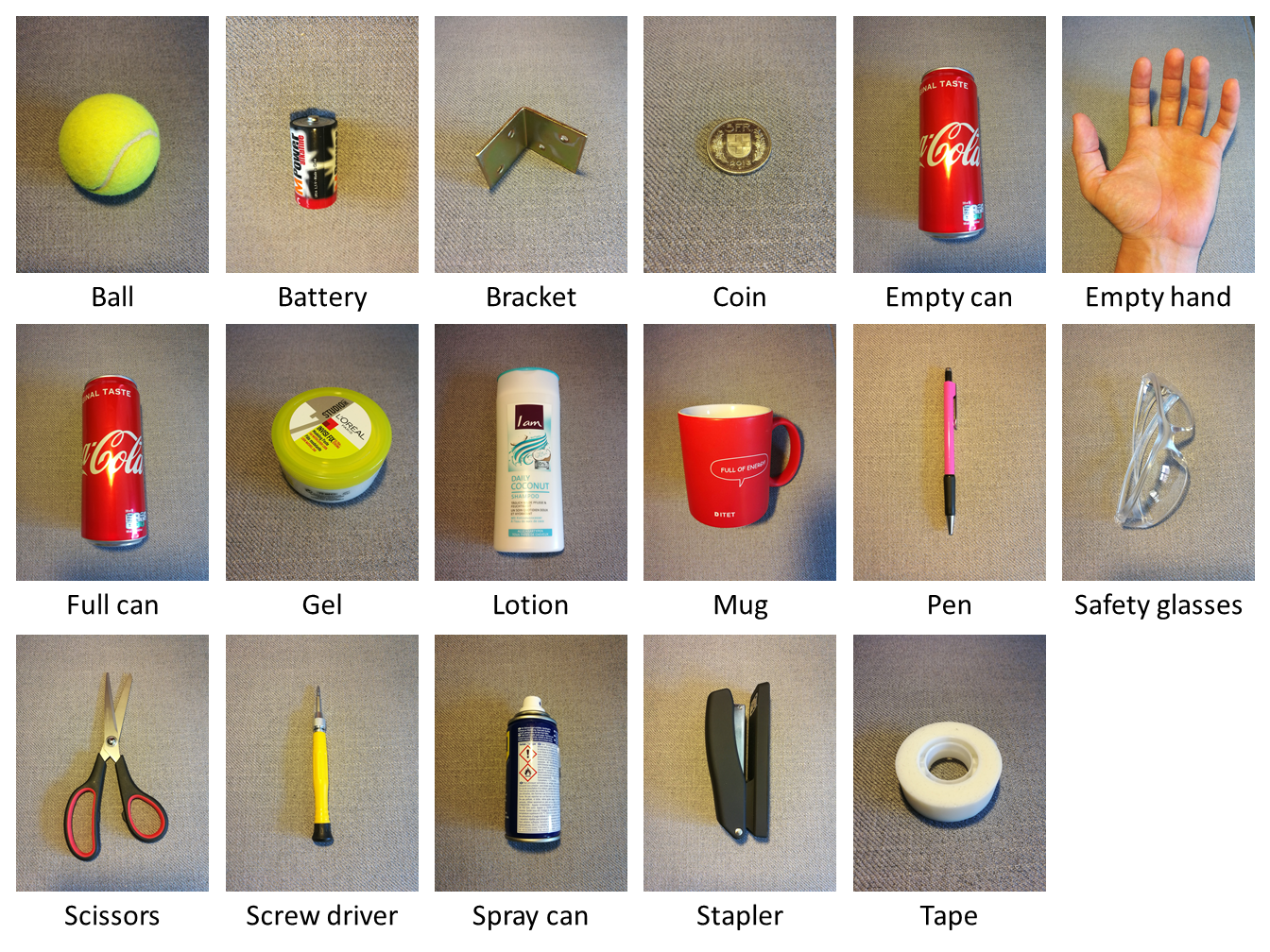}
  \caption{Pictures of the 16 objects and the empty hand used for the ETHZ-STAG dataset recording.}%
  \label{fig:dataset_objects}
\end{figure}

Following the analogous procedures of fabrication, we create our own \gls{stag} at ETH Zurich. 
A large dataset is collected with the \emph{data collection} modality described in Sec.~\ref{subsec:data_collection} in five recording sessions. Instead of synchronous video footage, synchronous accelerometer and gyroscope data are collected from an \gls{imu}.
\rebuttal{We select 16 objects that are easily found in our daily life and are as similar as possible to the ones used in~\cite{Sundaram2019}.
During every recording session, each object is manipulated for 40 seconds while continuously collecting tactile and \gls{imu} data at a rate of 100\,Hz.} 
\rebuttal{The manipulations are done in the same manner as in~\cite{Sundaram2019} to increase repeatability between recording sessions, i.e., the objects are manipulated by mostly pushing the sensor from the top onto the object lying on a table.}
The valid frames of actual object contact are marked by comparing to a threshold value for each taxel, as in~\cite{Sundaram2019}.
We construct these thresholds by finding the highest possible value that a taxel can assume using more than 20,000 additional empty-hand-frames of different hand poses without object contact.
Finally, a total of 340,000 frames are available \rebuttal{for 17 classes as shown in Fig.~\ref{fig:dataset_objects}}.

\subsection{Neural Network} \label{sec:cnn}
This section describes the model design, the training and the evaluation of the neural network used for processing the sensor data, and its embedded deployment on the \gls{mcu} for on-board processing.

\subsubsection{Convolutional neural network design}
\begin{figure*}
  \centering
  \includegraphics[width=.95\linewidth]{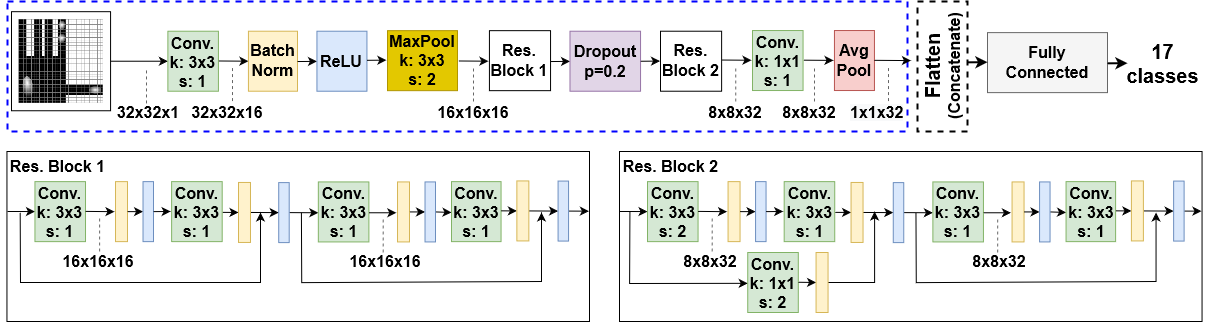}
  \caption{\rebuttal{The proposed \gls{cnn} based on ResNet-18. The feature maps size is HxWxC, with H and W being respectively the height and the width of the images and C the number of channels. k is the kernel size, s is the stride, and p is the dropout rate. The sensor fusion can be added by concatenating the extracted IMU features at the Flatten layer. The final model does not include the IMU data as explained in Sec.~\ref{subsec:sensorfusion}.}}
  \label{fig:adapted_cnn}
\end{figure*}

Because of the inherently spatial information present in the sensor, a \gls{cnn} architecture is chosen to process the tactile data. 
\rebuttal{As in~\cite{Sundaram2019}, we select a model architecture based on \mbox{ResNet-18}~\cite{he2016resnet18}, and redesign it by taking into consideration hardware resource constraints. Fig.~\ref{fig:adapted_cnn} depicts the proposed model architecture. 
It uses multiple stages of convolutions, including two residual blocks, to extract features from a single input frame. Finally, a fully connected layer is used for the classification.
Compared to~\cite{Sundaram2019}, we restrict the number of input frames to a single one directly reducing the latency.
Another adaptation is a four-fold reduction of convolution filters, i.e., 16 in the first convolutional layers and 32 in the second residual block. This decreases significantly the inference speed thanks to the reduced model complexity.}

\subsubsection{\rebuttal{Sensor fusion}}
To include the \gls{imu} data, a simple \gls{mlp} with one hidden layer consisting of 30 hidden units is added to extract the feature representations. The output layer of the \gls{mlp} consists of 3 neurons and is subsequently concatenated to the features of the tactile data before the final fully connected layer for the classification. This method is motivated by literature~\cite{Yunas2020sas,Trumble, Li2019}, in which the features of the individual sensors are first extracted and then concatenated. This additional branch featuring \gls{imu} data can be added or removed depending on the application.

\subsubsection{Training and validation}
The model is trained using PyTorch on GeForce GTX 1080 Ti GPUs with CUDA framework. 
\rebuttal{Two validation methodologies are employed: a) The full dataset is randomly split, and seven-fold \gls{cv} is performed. The classification accuracy is reported as the average of all folds. b) The dataset is split by session to consider the inter-session variability. We perform the leave-one-session-out \gls{cv}, i.e., one session is kept as validation set, while the remaining four are used for training. This procedure is repeated five times for each of the five sessions. The accuracy is reported as the average of all sessions.}
In both cases, the model is trained for 60 epochs with cross-entropy loss and Adam optimizer \rebuttal{with a batch size of 32 and a gradually decreasing learning rate as in~\cite{Sundaram2019} starting from 0.001. 
A pseudocode is shown in Algorithm~\ref{alg:pseudocode} and the source codes are open-source released.
}

\newcommand\mycommfont[1]{\footnotesize\ttfamily\textcolor{black}{#1}}
\SetCommentSty{mycommfont}
\begin{algorithm}[t]
\caption{\rebuttal{Neural network training and validation on our ETHZ-STAG dataset.}}\label{alg:pseudocode}
\KwData{Sensor data $X$}
\KwResult{Trained models}
$N_{folds}$ = 7 \tcp*[r]{7 folds for CV} 
$N_{sessions}$ = 5 \tcp*[r]{5 recording sessions}
\If{Random split}{
Shuffle $X$\;
Split $X$ into $N_{folds}$ training and validation sets\;
\For{$f$ from $1$ to $N_{folds}$}{
$X_{train,f}$ = training data from fold $f$\;
$X_{val,f}$ = validation data from fold $f$\;
Initialize model $m_f$\;
Train $m_f$ on $X_{train,f}$\;
Validate $m_f$ on $X_{val,f}$\;
}
}
\If{Inter-session}{
\For{$s$ from $1$ to $N_{sessions}$}{
$X_{val} = $ data from session $s$ for validation\;
$X_{train} = $ data from the remaining sessions for training\;
Initialize model $m_s$\;
Train $m_s$ on $X_{train,s}$\;
Validate $m_s$ on $X_{val,s}$\;
}
}
\end{algorithm}

\subsubsection{Emdedded implementation}
The X-CUBE-AI expansion package for STM32CubeMx is used for deploying the trained \gls{cnn} on the \gls{mcu}. For the embedded deployment, the model is trained by leveraging the full dataset. It is then saved in ONNX format and imported in STM32CubeMx. The tool generates an application template code, to which the user code to read the data can be added and the inference performed.

\new{
\subsection{Sensor Evaluations}\label{subsec:methods:sensoreval}

\rebuttal{With the purpose of evaluating the sensor properties, we perform two additional experiments.}
We fabricated a new control sensor as a square 16$\times$16 sensor array, as shown in \rebuttal{Fig.~\ref{fig:slip_experiment}}, and used it in the following experiments to investigate the physical properties of the sensor in a more controlled and repeatable manner:

\subsubsection{Sensor degradation}
The objective of this experiment was to determine, whether sensor degradation is caused by excessively moving, stressing and folding the sensor laminate --- as is the case for the sensor glove --- or if the degradation simply happens over time, even for a stationary sensor.
\rebuttal{We randomly choose an object (lotion) and place it on five different locations of the sensor area, four times within one week.}
This time frame is comparable to the last four recording sessions with the sensor glove, enabling the comparison of a strongly used sensor and a very lightly used sensor over roughly the same time period.

\subsubsection{\rebuttal{Slip evaluations}}
An extremely useful ability for robotic and prosthetic application is slip detection.
Tactile sensors have demonstrated the ability to detect object slip~\cite{Heyneman2016}, and even slip onset before the gripped object starts to move.
\rebuttal{Considering the importance of this property, we examine the sensor's reaction to slip or object motion.
In this experiment, the tactile sensor was lying flat on a table and an object was pulled across it at constant speed, to simulate a slipping event.}
Fig.~\ref{fig:slip_experiment} illustrates the experiment setup.

\rebuttal{The fabrication of the new sensor is justified by the following reasons: a) The purpose of the Smarthand sensor is to collect the dataset. After the dataset collection, the sensor’s properties have been affected by the usage, hence, it is not suitable for the additional evaluations. b) By fabricating a new sensor in the regular shape of a square, little strain is introduced during the fabrication preserving the sensor integrity yielding a more controlled experimental setup.
}

\begin{figure}[t]
  \centering
  \includegraphics[width=.75\linewidth]{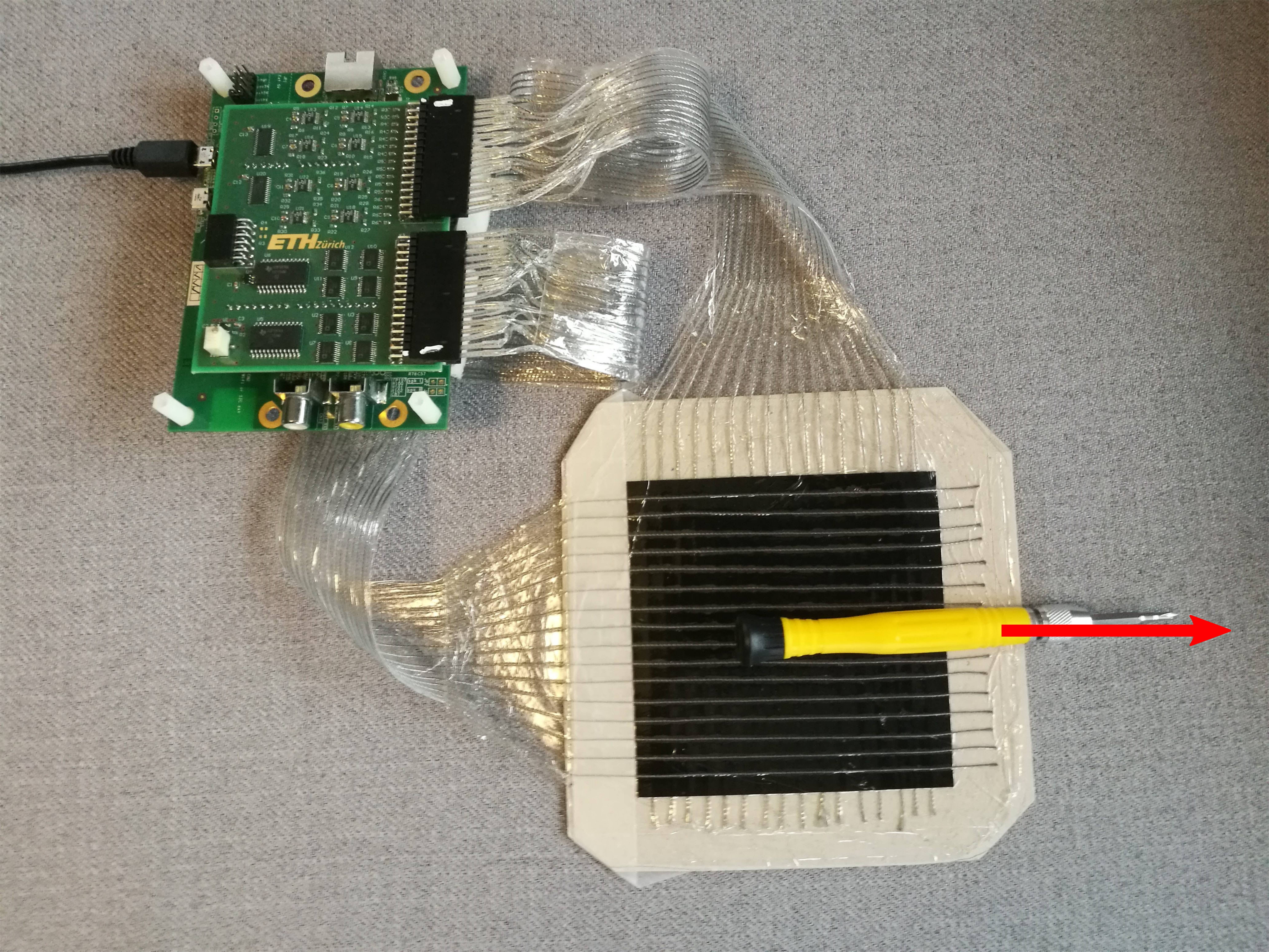}
  \caption{\rebuttal{Square sensor used as control sensor and a simple setup for simulating a slip event for a screwdriver.}}%
  \label{fig:slip_experiment}
\end{figure}

}
\section{Results}
\label{ch:results}

\rebuttal{In this section, we present the experimental results of object classification using our proposed system. Sec.~\ref{sec:neuralnet} presents the performance of our \gls{cnn}, its comparison to the previous work on the MIT-STAG dataset~\cite{Sundaram2019}, and the classification outcome on our dataset using the validation methodologies explained in Sec.~\ref{sec:cnn}. We discuss the sensor fusion approach in Sec.~\ref{subsec:sensorfusion} and report the outcome of the embedded implementation in terms of \glspl{macc}, memory usage, and inference time on the selected \gls{mcu} in Sec.~\ref{embedded_results}. The evaluations on the full system performance is discussed in Sec.~\ref{sys_perf_results}, while the additional sensor evaluations are reported in Sec.~\ref{subsec:sensoreval}. Finally, we discuss future works in Sec.~\ref{subsec:discussions}.}

\subsection{Neural network performance}\label{sec:neuralnet}

\definecolor{redisch}{RGB}{211,94,96}
\definecolor{greyisch}{RGB}{128,133,133}

\begin{figure}
\centering
  \resizebox{0.85\columnwidth}{!}{%
    \centering

\begin{tikzpicture}
\begin{axis}[
height=5cm,
width=9cm,
legend cell align={left},
legend style={at={(0.97,0.03)}, anchor=south east, draw=white!80.0!black},
tick align=outside,
tick pos=left,
x grid style={white!69.01960784313725!black},
xlabel={Number of frames},
xmajorgrids,
xmin=0, xmax=9,
xtick style={color=black},
xtick = {0, 1, ..., 9},
y grid style={white!69.01960784313725!black},
ylabel={Accuracy [\%]},
ymajorgrids,
ymin=0, ymax=100,
]
\addplot [thick, redisch, dotted, mark=*, mark size=1.5, mark options={solid}]
table [row sep=\\]{%
1 37.83\\
2 51.73\\
3 61.1\\
4 66.99\\
5 70.93\\
6 71.59\\
7 73.01\\
8 73.84\\
};
\addlegendentry{Cit.~\cite{Sundaram2019}}
\addplot [thick, greyisch, dotted, mark=*, mark size=1.5, mark options={solid}]
table[row sep=\\]{%
1 35.75\\
2 48.83\\
3 57.8\\
4 61.72\\
5 65.24\\
6 68.81\\
7 70.2\\
8 73.63\\
};
\addlegendentry{Ours}
\end{axis}

\end{tikzpicture}

  }

  \caption{\rebuttal{Comparison of the top-1 classification accuracy of our proposed model with the original network on MIT-STAG dataset~\cite{Sundaram2019} using different numbers of input frames.}}
  \label{fig:comparison_acc}
\end{figure}
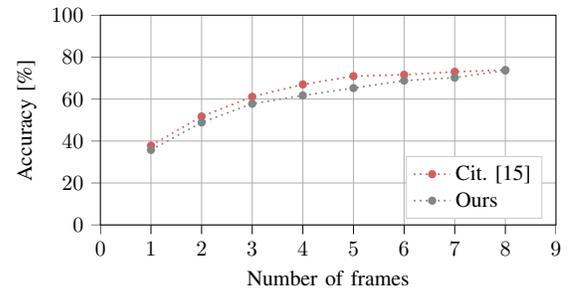

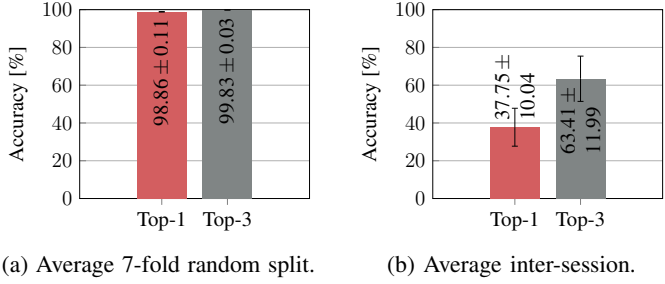
\begin{figure}
\centering
    \begin{subfigure}{0.47\columnwidth}
  \resizebox{\columnwidth}{!}{%
    \centering
\Large
\begin{tikzpicture}
\begin{axis}[
         ybar,
         xmin=0.8,xmax=2.2, 
         ymin=0,
         ymax=100,
         ymajorgrids,
         tick pos=left,
         y grid style={white!69.01960784313725!black},
         area legend,
         xtick={1.3,1.7}, 
         xticklabels={Top-1, Top-3}, 
         every axis plot/.append style={ 
          bar width=.3,
          bar shift=0pt,
          fill},
          ylabel={Accuracy [\%]},
         ]
        \addplot+ [
        draw=none,fill=redisch,
            error bars/.cd,
            error bar style={color=black},
                y dir=both,
                y explicit,
        ] coordinates {
            (1.3,98.86) +- (0,0.112)
        };
        \node[rotate=90,anchor=east] at (axis cs: 1.3, 98.86) {98.86\,$\pm$\,0.11};
        
        \addplot+ [
        draw=none,fill=greyisch,
            error bars/.cd,
            error bar style={color=black},
                y dir=both,
                y explicit
        ] coordinates {
            (1.7,99.83) +- (0,0.026)
        };
        \node[rotate=90,anchor=east] at (axis cs: 1.7, 99.83) {99.83\,$\pm$\,0.03};
       \end{axis}
            \end{tikzpicture}
            
}

    \caption{\rebuttal{Average 7-fold random split}.}
  \label{fig:randomsplit}
\end{subfigure}
\hfill
\begin{subfigure}{0.47\columnwidth}
  \resizebox{\columnwidth}{!}{%
    \centering
\Large
\begin{tikzpicture}
\begin{axis}[
         ybar,
         xmin=0.8,xmax=2.2, 
         ymin=0,
         ymax=100,
         ymajorgrids,
         tick pos=left,
         y grid style={white!69.01960784313725!black},
         area legend,
         xtick={1.3,1.7}, 
         xticklabels={Top-1, Top-3}, 
         every axis plot/.append style={ 
          bar width=.3,
          bar shift=0pt,
          fill},
          ylabel={Accuracy [\%]},
         ]
         \addplot+ [
        draw=none,fill=redisch,
            error bars/.cd,
            error bar style={color=black},
                y dir=both,
                y explicit
        ] coordinates {
            (1.3,37.75) +- (0,10.04)
        };
         \node[rotate=90,anchor=west, text width=1.6cm] at (axis cs: 1.3, 37.75) {37.75\,$\pm$ 10.04}; 
         
         \addplot+ [
        draw=none,fill=greyisch,
            error bars/.cd,
            error bar style={color=black},
                y dir=both,
                y explicit
        ] coordinates {
            (1.7,63.41) +- (0,11.99)
        };
         \node[rotate=90,anchor=east, text width=1.6cm] at (axis cs: 1.7, 63.41) {63.41\,$\pm$ 11.99};
       \end{axis}
            \end{tikzpicture}
            
}

    \caption{\rebuttal{Average inter-session}.}
  \label{fig:intersessionavg}
\end{subfigure}
\caption{\rebuttal{Classification accuracy on ETHZ-STAG dataset.}}
    \label{fig:accuracy}
\end{figure}

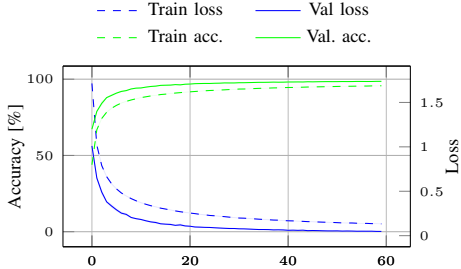
\begin{figure}[t]
    \centering
\begin{tikzpicture}
\begin{axis}[
no markers,
width=0.7\linewidth,
height=0.45\linewidth,
        ticklabel style={font={\tiny}},
         y grid style={white!69.01960784313725!black},
          ylabel={\scriptsize Loss},
          ylabel shift = -6 pt,
          axis y line* = right
         ]
\addplot+ [blue,dashed] table[x=epochs,y=trainloss_avg, col sep=comma] {figures/data/np44225_baseline_avg_std.csv};

\addplot+ [name path=upper,draw=none] table[x=epochs,y expr=\thisrow{trainloss_avg}+\thisrow{trainloss_std},col sep=comma] {figures/data/np44225_baseline_avg_std.csv};
\addplot+ [name path=lower,draw=none] table[x=epochs,y expr=\thisrow{trainloss_avg}-\thisrow{trainloss_std},col sep=comma] {figures/data/np44225_baseline_avg_std.csv};
\addplot+ [fill=blue!10] fill between[of=upper and lower];

\addplot+ [blue] table[x=epochs,y=testloss_avg, col sep=comma] {figures/data/np44225_baseline_avg_std.csv};

\addplot+ [name path=upper,draw=none] table[x=epochs,y expr=\thisrow{testloss_avg}+\thisrow{testloss_std},col sep=comma] {figures/data/np44225_baseline_avg_std.csv};
\addplot+ [name path=lower,draw=none] table[x=epochs,y expr=\thisrow{testloss_avg}-\thisrow{testloss_std},col sep=comma] {figures/data/np44225_baseline_avg_std.csv};
\addplot+ [fill=blue!10] fill between[of=upper and lower];
       \end{axis}
\begin{axis}[
no markers,
width=0.7\linewidth,
height=0.45\linewidth,
        ticklabel style={font={\tiny}},
        grid = both,
        axis y line* = left,
         y grid style={white!69.01960784313725!black},
          ylabel={\scriptsize Accuracy [\%]},
          ylabel shift = -6 pt,
                legend style={
        at={(0.5, 1.05)},
        anchor=south,
        draw=none,
        /tikz/column 2/.style={column sep=10pt},
        /tikz/column 4/.style={column sep=10pt},
        font={\scriptsize}
      },
      legend columns = 2
         ]
         
         \addplot+ [blue,dashed] coordinates {(0, 0)};
         
         \addplot+ [blue] coordinates {(0, 0)};
         
\addplot+ [green,dashed] table[x=epochs,y=trainprec_avg, col sep=comma] {figures/data/np44225_baseline_avg_std.csv};

\addplot+ [name path=upper,draw=none,forget plot] table[x=epochs,y expr=\thisrow{trainprec_avg}+\thisrow{trainprec_std},col sep=comma] {figures/data/np44225_baseline_avg_std.csv};
\addplot+ [name path=lower,draw=none,forget plot] table[x=epochs,y expr=\thisrow{trainprec_avg}-\thisrow{trainprec_std},col sep=comma] {figures/data/np44225_baseline_avg_std.csv};
\addplot+ [fill=green!10,forget plot] fill between[of=upper and lower];

\addplot+ [green] table[x=epochs,y=testprec_avg, col sep=comma] {figures/data/np44225_baseline_avg_std.csv};

\addplot+ [name path=upper,draw=none,forget plot] table[x=epochs,y expr=\thisrow{testprec_avg}+\thisrow{testprec_std},col sep=comma] {figures/data/np44225_baseline_avg_std.csv};
\addplot+ [name path=lower,draw=none,forget plot] table[x=epochs,y expr=\thisrow{testprec_avg}-\thisrow{testprec_std},col sep=comma] {figures/data/np44225_baseline_avg_std.csv};
\addplot+ [fill=green!10,forget plot] fill between[of=upper and lower];
\legend{Train loss, Val loss, Train acc., Val. acc.};
       \end{axis}
            \end{tikzpicture}

\caption{\rebuttal{Average \gls{cv} learning curves on ETHZ-STAG dataset.}}
    \label{fig:training_curves}
\end{figure}

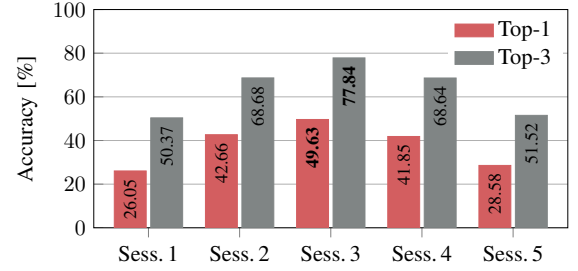
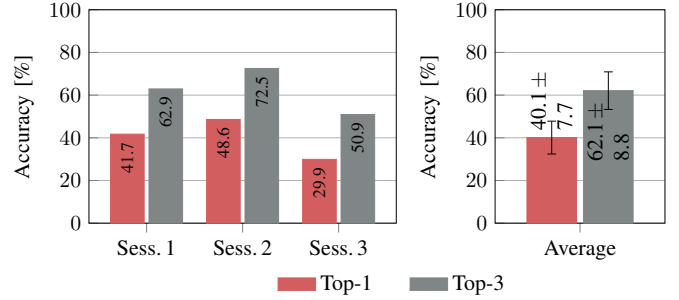
\begin{figure}[t]
\centering

\begin{subfigure}{\columnwidth}
\centering
\resizebox{0.85\columnwidth}{!}{%
\begin{tikzpicture}
\begin{axis}[
width=9cm,
height=5cm,
         ybar,
         xmin=0.4,xmax=5.6, 
         ymin=0,
         ymax=100,
         ymajorgrids,
         tick pos=left,
         y grid style={white!69.01960784313725!black},
         area legend,
         xtick={1,2,3,4,5}, 
         xticklabels={Sess.\,1, Sess.\,2, Sess.\,3, Sess.\,4, Sess.\,5}, 
         every axis plot/.append style={ 
          bar width=.35,
          bar shift=0pt,
          fill},
          ylabel={Accuracy [\%]},
          legend style={at={(0.99,0.99)},anchor=north east,draw=none, 
                },
         ]
         
        \addplot[draw=redisch,fill=redisch] coordinates {(0.8,26.052)};
        \node[rotate=90,anchor=east] at (axis cs: 0.8, 26.052) {\footnotesize 26.05};
        \addlegendentry{Top-1}
        \addplot[draw=greyisch,fill=greyisch] coordinates {(1.2,50.370)};
        \node[rotate=90,anchor=east] at (axis cs: 1.2, 50.370) {\footnotesize 50.37};
        \addlegendentry{Top-3}
        
        \addplot[draw=redisch,fill=redisch] coordinates {(1.8,42.656)};
        \node[rotate=90,anchor=east] at (axis cs: 1.8, 42.656) {\footnotesize 42.66};
        \addplot[draw=greyisch,fill=greyisch] coordinates {(2.2,68.677)};
        \node[rotate=90,anchor=east] at (axis cs: 2.2, 68.677) {\footnotesize 68.68};
        
        \addplot[draw=redisch,fill=redisch] coordinates {(2.8,49.632)};
        \node[rotate=90,anchor=east] at (axis cs: 2.8, 49.632) {\footnotesize \textbf{49.63}};
        \addplot[draw=greyisch,fill=greyisch] coordinates {(3.2,77.842)};
        \node[rotate=90,anchor=east] at (axis cs: 3.2, 77.842) {\footnotesize \textbf{77.84}};
         
        \addplot[draw=redisch,fill=redisch] coordinates {(3.8,41.845)};
        \node[rotate=90,anchor=east] at (axis cs: 3.8, 41.845) {\footnotesize 41.85};
        \addplot[draw=greyisch,fill=greyisch] coordinates {(4.2,68.641)};
        \node[rotate=90,anchor=east] at (axis cs: 4.2, 68.641) {\footnotesize 68.64};
        
        \addplot[draw=redisch,fill=redisch] coordinates {(4.8,28.583)};
        \node[rotate=90,anchor=east] at (axis cs: 4.8, 28.583) {\footnotesize 28.58};
        \addplot[draw=greyisch,fill=greyisch] coordinates {(5.2,51.523)};
        \node[rotate=90,anchor=east] at (axis cs: 5.2, 51.523) {\footnotesize 51.52};
         
       \end{axis}
            \end{tikzpicture}
}
\caption{ETHZ-STAG dataset.}
\label{fig:intersession:ethzstag}
\end{subfigure}
\vfill
\begin{subfigure}{\columnwidth}
\centering
\resizebox{\columnwidth}{!}{%
\begin{tikzpicture}
\begin{axis}[
width=6.5cm,
height=5cm,
xshift=-4.5cm,
         ybar,
         xmin=0.4,xmax=3.6, 
         ymin=0,
         ymax=100,
         ymajorgrids,
         tick pos=left,
         y grid style={white!69.01960784313725!black},
         area legend,
         xtick={1,2,3}, 
         xticklabels={Sess.\,1, Sess.\,2, Sess.\,3}, 
         every axis plot/.append style={ 
          bar width=.35,
          bar shift=0pt,
          fill},
          ylabel={Accuracy [\%]},
          legend style={at={(0.99,-0.2)},anchor=north,draw=none, 
          /tikz/column 2/.style={
                column sep=15pt,}
                },
          legend columns=2,
         ]
         
        \addplot[draw=redisch,fill=redisch] coordinates {(0.8,41.7)};
        \node[rotate=90,anchor=east] at (axis cs: 0.8, 41.7) {\footnotesize 41.7};
        \addlegendentry{Top-1}
        \addplot[draw=greyisch,fill=greyisch] coordinates {(1.2,62.9)};
        \node[rotate=90,anchor=east] at (axis cs: 1.2, 62.9) {\footnotesize 62.9};
        \addlegendentry{Top-3}
        
        \addplot[draw=redisch,fill=redisch] coordinates {(1.8,48.6)};
        \node[rotate=90,anchor=east] at (axis cs: 1.8, 48.6) {\footnotesize 48.6};
        \addplot[draw=greyisch,fill=greyisch] coordinates {(2.2,72.5)};
        \node[rotate=90,anchor=east] at (axis cs: 2.2, 72.5) {\footnotesize 72.5};
        
        \addplot[draw=redisch,fill=redisch] coordinates {(2.8,29.9)};
        \node[rotate=90,anchor=east] at (axis cs: 2.8, 29.9) {\footnotesize 29.9};
        \addplot[draw=greyisch,fill=greyisch] coordinates {(3.2,50.9)};
        \node[rotate=90,anchor=east] at (axis cs: 3.2, 50.9) {\footnotesize 50.9};
        
       \end{axis}
       
\begin{axis}[
width=4.3cm,
height=5cm,
xshift=2cm,
         ybar,
         xmin=0.4,xmax=1.6, 
         ymin=0,
         ymax=100,
         ymajorgrids,
         tick pos=left,
         y grid style={white!69.01960784313725!black},
         area legend,
         xtick={1}, 
         xticklabels={Average}, 
         every axis plot/.append style={ 
          bar width=.35,
          bar shift=0pt,
          fill},
          ylabel={Accuracy [\%]},
          legend style={at={(0.99,0.99)},anchor=north east,draw=none, 
                },
         ]
        
         \addplot+ [draw=redisch,fill=redisch, error bars/.cd, error bar style={color=black}, y dir=both, y explicit] coordinates 
         {(0.8,40.1) +- (0,7.7)};
         \node[rotate=90,anchor=west, text width=1.4cm] at (axis cs: 0.8, 40.1) {40.1\,$\pm$ 7.7};
         \addplot+ [draw=greyisch,fill=greyisch, error bars/.cd, error bar style={color=black}, y dir=both, y explicit] coordinates 
         {(1.2, 62.1) +- (0,8.8)};
         \node[rotate=90,anchor=east, text width=1cm] at (axis cs: 1.2, 62.1) {62.1\,$\pm$ 8.8};
        
       \end{axis}
            \end{tikzpicture}
}
\caption{MIT-STAG network and dataset.}
\label{fig:intersession:mitstag}
\end{subfigure}
  \caption{\rebuttal{Inter-session accuracy. Top 1 and top 3 accuracy when using one entire recording session for testing and the remaining sessions for training.}}%
  \label{fig:intersession}
\end{figure}

\rebuttal{First of all, we compare the accuracy of our proposed model with the original network using the MIT-STAG dataset comprising of 26 objects~\cite{Sundaram2019}. The authors randomly select N frames from each recording of several minutes as the input to the network. We reproduce their results with their \gls{cnn} and compare them with our proposed model trained on the same dataset with the same training and validation methodology~\cite{Sundaram2019}.} As shown in Fig.~\ref{fig:comparison_acc}, despite the significantly smaller size of our model, the accuracy is comparable, especially for N = 8, where the accuracy drop is only 0.21\%. However, N = 1 would introduce less computational burden for a real-time application scenario yielding lower latency. Again, our model performs comparably to the original one, with less than 2\% accuracy drop with \mbox{N = 1}.

The second step is to evaluate the model performance on our dataset collected by the \gls{stag} fabricated in-house and integrated into our proposed system.
With our dataset, we consider only N = 1 and take all the valid frames instead of randomly selecting subsets of valid frames from each recording. This reduces the several-minute latency of previous work and enables the real-time response of the entire system. 
Fig.~\ref{fig:randomsplit} shows the results for the seven-fold cross-validation with random splitting on the whole dataset without considering inter-session variability.
The top-1 and top-3 accuracy values, i.e., the correct class is the one with the highest predicted probability, and the correct class is among the ones with the three highest probabilities, are respectively 98.86\% and 99.83\%.
\rebuttal{Fig.~\ref{fig:training_curves} depicts the learning curves averaged over the 7 folds. We observe that both training and validation accuracy and loss converge to a plateau after 60 epochs.
This demonstrates that the model is able to learn the full data distribution.} 

\rebuttal{We then proceed with inter-session validation, i.e., the model is trained on the data from four sessions and is validated on the data from the remaining session. The top-1 and the top-3 average accuracy over the five sessions are shown in Fig.~\ref{fig:intersessionavg}.
The drop in accuracy is expected due to the more challenging problem caused by the inter-session variability, where the model does not generalize very well to the unseen sessions. It is a common phenomenon observed in biomedical applications~\cite{wang2020accurate}. In the following paragraphs, we analyze more in depth the inter-session variability, which is likely caused by the differences in object interaction and in the wear of the glove between sessions. Another possible explanation is the sensor degradation analyzed in Sec.~\ref{subsec:sensoreval}.}

\new{
\rebuttal{
In Fig.~\ref{fig:intersession:ethzstag}, we unwrap the results from the inter-session validation and demonstrate the accuracy on each session. We can observe that the performance with the leave-one-session-out cross-validation strongly depends on the specific session used as validation set. The best top-1 (49.63\%) and top-3 (77.84\%) accuracy results are reached by using the session 3 as validation set, while training the network on the remaining sessions.
The related work in~\cite{Sundaram2019} did not present any inter-session validation results, however, for comparison, we reproduce their network on the MIT-STAG dataset using the same inter-session validation methodology used here. As can be seen in Fig.~\ref{fig:intersession:mitstag}, similar results are obtained, proving that the accuracy drop is not cause by our dataset and network, but it is a result of the inter-session variability.

Next, we investigate the confusion matrices of different sessions to gain a better understanding of the problems faced by the machine learning model.}
The confusion matrix in Fig.~\ref{fig:sess3_cm} reveals that there are some objects that are classified well, while others are much `harder' to classify.
Additionally, similar objects, such as a screwdriver and a pen, and objects with a non-uniform shape, such as scissors and safety glasses, are more likely to be confused with each other.
This is intuitive and can be largely attributed to the fact that recording sessions are not perfectly repeatable, because of small differences in interaction or in wearing the sensor glove.
\begin{figure}
  \centering
  \includegraphics[width=0.95\linewidth]{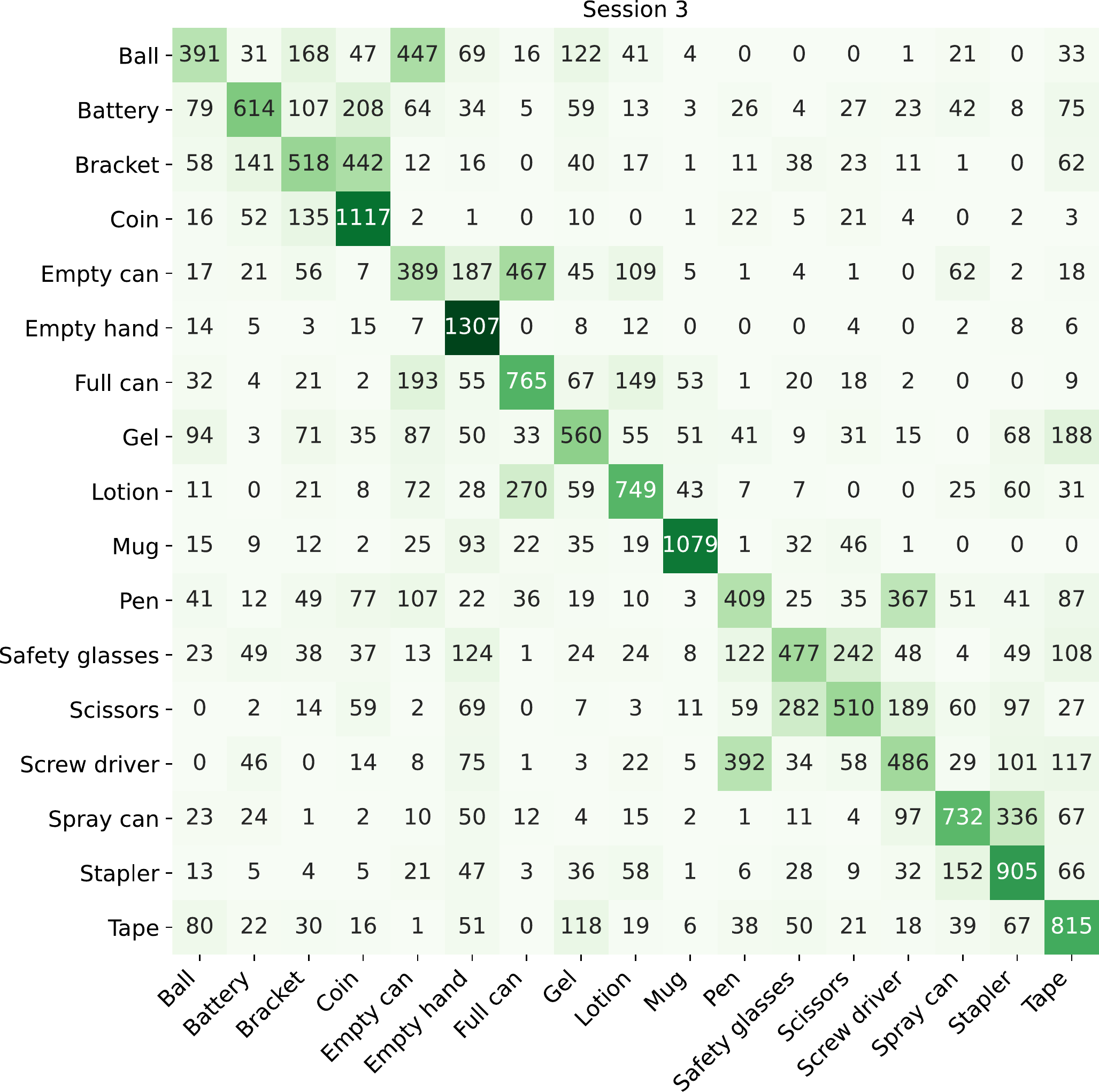}
  \caption{Confusion matrix when using recording session 3 as test data. The rows are the true class and the columns are the predicted ones.}%
  \label{fig:sess3_cm}
\end{figure}
However, this explanation does not apply to all the sessions. Looking at the confusion matrix of the worst performing session in Fig.~\ref{fig:sess1_cm}, some classes are classified well, but most others are not much better than random guessing.
The main source of  `confusion' does not seem to be similarity between objects, and it is not obvious from the confusion matrix what the reason for such a low accuracy is.
\rebuttal{Another possible explanation is presented in Sec.~\ref{subsec:sensoreval} with the additional evaluations on sensor degradation.}
\begin{figure}
  \centering
  \includegraphics[width=0.95\linewidth]{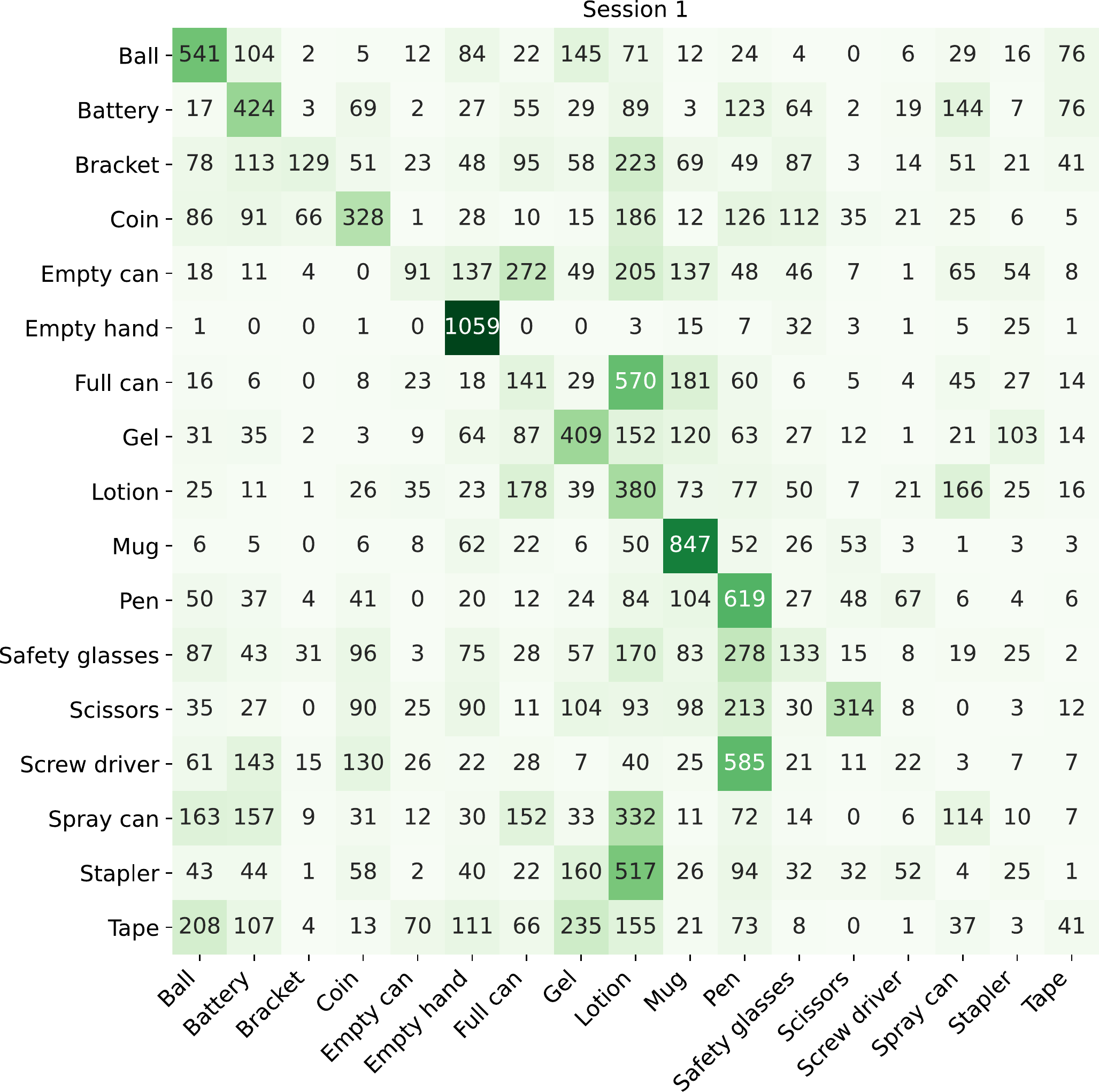}
  \caption{Confusion matrix when using recording session 1 as test data. The rows are the true class and the columns are the predicted ones.}%
  \label{fig:sess1_cm}
\end{figure}

}

\subsection{Sensor fusion}\label{subsec:sensorfusion}

The \gls{imu} orientation in the form of Euler angles is directly fed into the \gls{mlp}, or only accelerometer or gyroscope data is used. 
\rebuttal{No configuration has provided any improvement over the model using only tactile data.
The usage of \gls{imu} data in this application scenario is not very meaningful, as all the objects were mostly manipulated from the top.
Hence, for the final evaluations presented in this paper, we exclude the \gls{imu} sensor.}
In a real application scenario with robotic hands, the information about end-effector orientation is crucial and needs to be included at various stages of the processing loop.
Accordingly, the integration of the \gls{imu} is a step towards a usable and practical system.
\rebuttal{We discuss future works in Sec.~\ref{subsec:discussions} that can potentially benefit from the inclusion of the \gls{imu} sensor.}

\subsection{Embedded implementation}\label{embedded_results}

We subsequently proceed with network deployment on the \gls{mcu}. Here the main advantage of our proposed model becomes evident.
In fact, the proposed network in~\cite{Sundaram2019} is too big for the selected \gls{mcu} making it impossible to be embedded onboard. Table~\ref{tab:network_comparison} reports the comparison between the two networks. 
Note that the inference time reported for the original network is obtained by reducing the input convolution filters from 64 to 54 since it is impossible to deploy the full network due to memory constraints.
Our model requires one order of magnitude less memory, making it possible to be deployed. The number of computations is reduced by 15.6$\times$ in terms of \glspl{macc}, yielding more than 12$\times$ speedup for the inference time. This effectively enables a system with real-time response. 

\begin{table}[b]
  \caption{\rebuttal{Comparison of neural networks embedded on STM32F769NI MCU using CUBE-AI.}}
  \label{tab:network_comparison}
  \centering\begin{tabular}{@{}lrrrr@{}} \toprule
  \textbf{Project} & \textbf{\acrshort{macc}} & \textbf{Flash} & \textbf{\acrshort{ram}} & \textbf{Inference time}\\ \midrule
  Cit.~\cite{Sundaram2019} & 73.3 M & 2.77 MB & 196 kB & >1.2 s \\
  This work & 4.7 M & 177 kB & 52 kB & 100 ms \\ \bottomrule
  \end{tabular}
\end{table}

\new{
\subsection{System performance}\label{sys_perf_results}

As explained in Sec.~\ref{subsec:data_collection}, there are different applications which have different processing throughput, according to how their base timer was programmed.
Table~\ref{tab:throughput} summarizes the throughput and system clock of each application.
Compared to the data collection in~\cite{Sundaram2019}, our system has a throughput of 100 \gls{fps}, which is 13.7$\times$ higher (see Table~\ref{tab:front_end_comparison}). 
The usable signal range was increased by adjusting the analog circuit response, and additionally to increasing the signal range, the \gls{adc} reference voltage was reduced from 5 V to 3.3 V.
Thanks to the high data acquisition rate of the analog front-end, significantly more frames can be collected in a shorter amount of time.
In fact, five data collection sessions were performed in this work, compared to the three sessions of~\cite{Sundaram2019}. This allowed a more extensive analysis on the inter-session variability.

\begin{table}[b]
  \caption{Processing throughput.}
  \label{tab:throughput}
  \centering\begin{tabular}{@{}lrrr@{}} \toprule
  \textbf{Application} & Data collection & \acrshort{gui} & Complete system\\ \midrule
  \textbf{Throughput [Hz]} & 100 & 10 & 8\\
  \textbf{System clock [MHz]} & 144 & 144 & 216\\ \bottomrule
  \end{tabular}
\end{table}
\begin{table}[b]
  \caption{\rebuttal{Comparison of our improved \gls{afe} based on~\cite{Sundaram2019}.}}
  \label{tab:front_end_comparison}
  \centering\begin{tabular}{@{}lrrrr@{}} \toprule
  \textbf{Project} & \textbf{Throughput} & \textbf{Power} & \textbf{Resolution} & \textbf{Range}\\ \midrule
  Cit.~\cite{Sundaram2019} & 7.3 fps & -- & 10 bit & 0.7 V \\
  This work & 100 fps & max. 52 mW & 12 bit & 1.2 V \\ \bottomrule
  \end{tabular}
\end{table}

Next, we present a detailed characterization of the system's power consumption, showing its feasibility and simultaneously underscoring potential improvements.
All power measurements were made while running the complete system application, implementing data acquisition and neural network inference.
This is the most relevant firmware for a possible real-world application. 
The power characterization is performed with a Keysight N6705C power analyzer.
The subsystems are the \gls{imu}, the \gls{mcu}, the readout circuit, and the discovery board excluding the \gls{mcu}.
With the exception of the discovery board and \gls{mcu}, the subsystems are fully independent physical boards, making a distinction of the individual power domains straightforward.

The \gls{imu} and \gls{mcu} were supplied with 3.3V while the discovery board and the readout circuit were supplied with 5V.
Additionally to shorting all the negative outputs of the power analyzer, the ground across all subsystems was shorted.
Also, the discovery board was only supplied to ensure proper functioning of the application but will be excluded from the analysis.
The power consumption of the actively and continuously running system is shown in Table~\ref{tab:power}.
The consumption of the \gls{imu} (MPU-9250) agrees with the data sheet, considering that all nine axes are active at full speed and that it is supplied with 3.3V.
The same is true for the microcontroller (STM32F769NI), which runs at its maximum clock speed of 216MHz.
Regarding the readout circuit, a power consumption of 52mW is reasonable because it uses a \gls{ldo} regulator to generate a reference voltage of 1.2V from an input voltage of 5V.

We further characterize the low-power states of all components. 
The system first collects and processes data for ten seconds and then puts all subsystems into a low-power state, from which it can be woken at arbitrary moments by pressing a button on the discovery board.
Fig.~\ref{fig:i_imu_run}, Fig.~\ref{fig:i_mcu} and Fig.~\ref{fig:i_readoutCircuit} display current traces that were acquired with the power analyzer during the low-power application.
The periodicity of the \gls{mcu} and readout circuit trace in the active sections clearly corresponds to the frequency of the processing timer.
Currently there is no possibility to completely turn off the readout circuit.
This leads to the observed behavior during the standby phases in Fig.~\ref{fig:i_readoutCircuit}.
It is caused by the fact that pins of the \gls{mcu} are left floating when it enters standby mode.
In future versions, a digital power switch can easily be added to the readout circuit to reduce the current consumption to a minimum during standby.
A conservative estimate is that such a power switch will dissipate no more than 1\textmu W.

\begin{figure}[t]
  \centering
  \includegraphics[width=\linewidth]{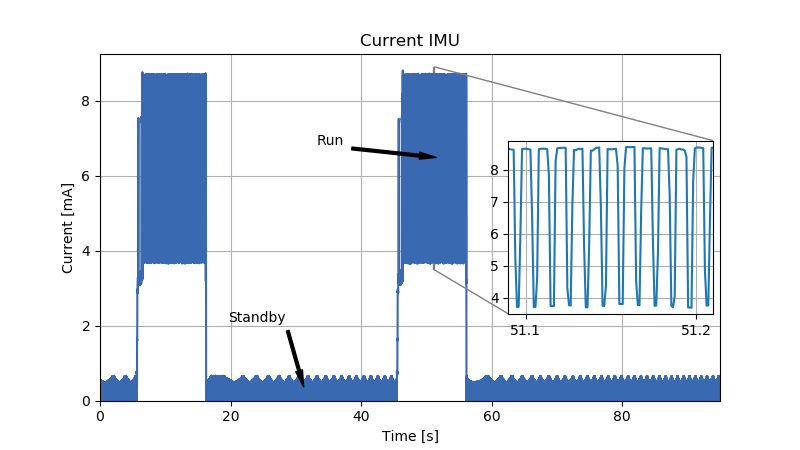}
  \caption{Measured current of the \gls{imu} during a simulated low-power application.}%
  \label{fig:i_imu_run}
\end{figure}
\begin{figure}[t]
  \centering
  \includegraphics[width=\linewidth]{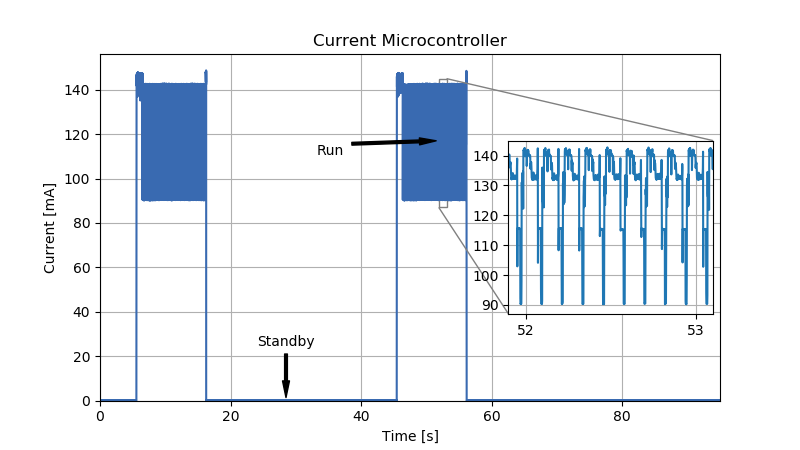}
  \caption{Measured current of the \gls{mcu} during a simulated low-power application.}%
  \label{fig:i_mcu}
\end{figure}
\begin{figure}[t]
  \centering
  \includegraphics[width=\linewidth]{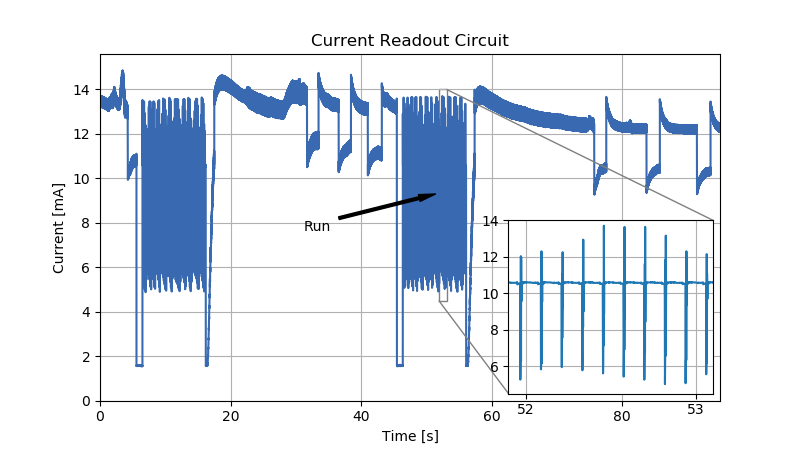}
  \caption{Measured current of the readout circuit during a simulated low-power application.}%
  \label{fig:i_readoutCircuit}
\end{figure}

By utilizing a special low-power accelerometer-only mode, the \gls{imu} is not in complete standby but samples accelerometer data at a configurable rate.
Fig.~\ref{fig:i_imu_standby} displays a closer look at the standby section, revealing this sampling behavior.
The output data rate was configured to 31.25Hz which is readily recognizable from the plot.
A feature called \gls{wom} enables the \gls{imu} to send an interrupt if any accelerometer axis exceeds a programmable threshold, which could be used in a real application to wake the \gls{mcu} if object contact is detected.

\begin{figure}[t]
  \centering
  \includegraphics[width=\linewidth]{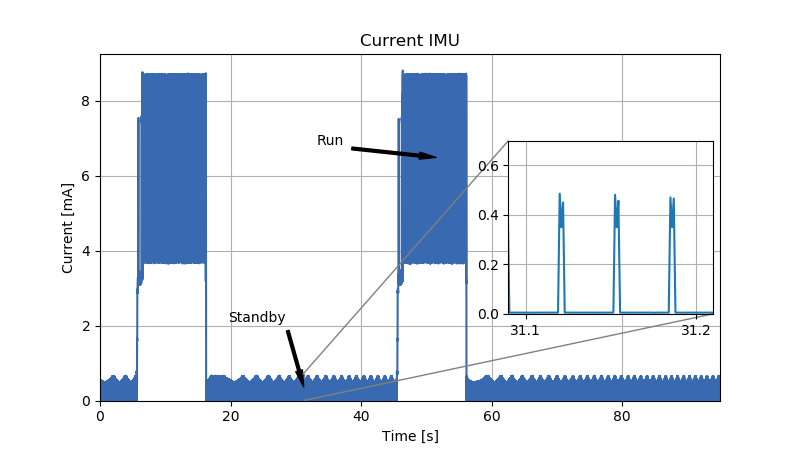}
  \caption{Measured current of the \gls{imu} during a simulated low-power application.}%
  \label{fig:i_imu_standby}
\end{figure}

Measuring the power consumption during standby gives the possibility of realistically predicting the lifetime of the embedded system.
A summary of relevant numbers is given in Table~\ref{tab:power}.
Again, the measured current draw of \gls{imu} and \gls{mcu} was compared to the numbers given in the datasheet.
Both supply currents are slightly higher than specified.
For the inertial measurement unit this comes from the fact that it is supplied with 3.3V, while the numbers in the datasheet correspond to a supply voltage of 2.5V.
In the case of the microcontroller, the current draw is a bit more than twice the specified value.
Setting all pins of the \gls{mcu} to analog before entering standby mode will most likely reduce the current to the value given in the datasheet.

\begin{table}[b]
  \caption{Power consumption.}
  \label{tab:power}
  \centering\begin{tabular}{@{}lrrrr@{}} \toprule
  \textbf{Subsystem} & \gls{imu} & \gls{mcu} & Readout circuit & Total\\ \midrule
  \textbf{Supply voltage} & 3.3V & 3.3V & 5V & -- \\
  \textbf{Power run} & 23mW & 430mW & 52mW & 505mW \\
  \textbf{Power standby} & 144µW & 40µW & 1µW & 185µW \\ \bottomrule
  \end{tabular}
\end{table}

\vlad{We note the time when the system is in run mode and standby mode as $t\sb{ON}$ and $t\sb{OFF}$, respectively.
Furthermore, we define as duty-cycle the time spent in run mode divided by the whole period, and therefore $DC = \frac{t\sb{ON}}{t\sb{ON} + t\sb{OFF}}$ \cite{mayer2020smart}.
As the system consumes $P\sb{ON} = \SI{505}{\milli \watt}$ during the run mode and $P\sb{OFF} = \SI{0.185}{\milli \watt}$ during standby, the average power during a whole period (i.e., $t\sb{ON}$ + $t\sb{OFF}$) can be calculated using Equation \ref{eq:avg_power}.

\begin{equation} \label{eq:avg_power}
  P_{avg} = (1 - DC) \cdot P\sb{OFF} + DC \cdot P\sb{ON}
\end{equation}
}

Assuming a duty cycle of 10\% and an operating time of 20 hours per day, an energy of $E = P_{avg} \cdot 20h = (0.9 \cdot 0.185mW + 0.1 \cdot 505mW) \cdot 20h \approx 1Wh$ is required.
A typical smart phone battery has a capacity of more than 10Wh. This means that the current prototype could last 20 hours on a battery with only one tenth the size of a smartphone battery, even when assuming the worst-case power consumption.

To show that power optimization of the system is still possible and advisable, a small experiment was conducted by supplying the readout circuit with 4.5V instead of 5V.
The resulting power consumption of the readout circuit was 17mW --- a 0.5V supply voltage reduction thus led to a three times lower power consumption.
The main reason for this large influence of the supply voltage is the \gls{ldo} regulator that is used to generate a reference voltage of 1.2V.
By decreasing the difference between input and output voltage of the \gls{ldo} regulator, the power dissipation in this component is directly reduced.
Furthermore, the power consumption of other \glspl{ic} on the \gls{pcb} is reduced.

}

\new{
\subsection{Sensor evaluations}\label{subsec:sensoreval}

\begin{figure}[b]
  \centering
  \subfloat[S1: 1.0]{\includegraphics[width=0.2\linewidth]{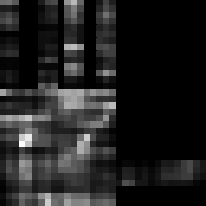}}\hfill
  \subfloat[S2: 0.74]{\includegraphics[width=0.2\linewidth]{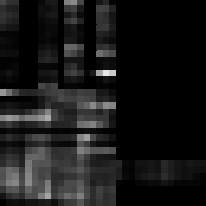}}\hfill
  \subfloat[S3: 0.69]{\includegraphics[width=0.2\linewidth]{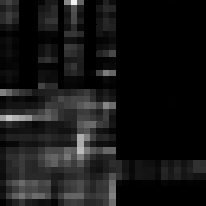}}\hfill
  \subfloat[S4: 0.58]{\includegraphics[width=0.2\linewidth]{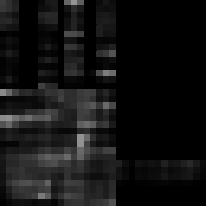}}\hfill
  \subfloat[S5: 0.45]{\includegraphics[width=0.2\linewidth]{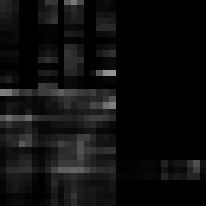}}
  \caption{\rebuttal{Average of all contact-frames from each recording session for class 13 (screw driver). From (a) to (e) are respectively session 1 to 5. The numbers represent the relative mean response of each session compared to session 1.}}
  \label{fig:comparison_class13}
\end{figure}

\definecolor{redisch}{RGB}{211,94,96}
\definecolor{greyisch}{RGB}{128,133,133}



\begin{figure}[t]
\centering
  \resizebox{0.85\columnwidth}{!}{%
    \centering

\begin{tikzpicture}
\begin{axis}[
height=5cm,
width=9cm,
legend cell align={left},
legend style={at={(0.97,0.03)}, anchor=south east, draw=white!80.0!black},
tick align=outside,
tick pos=left,
x grid style={white!69.01960784313725!black},
xlabel={Session},
xmajorgrids,
xmin=0, xmax=6,
xtick style={color=black},
xtick = {0, 1, ..., 6},
y grid style={white!69.01960784313725!black},
ylabel={Relative mean response},
ymajorgrids,
ymin=0.0, ymax=1.1,
ytick = {0.0, 0.2, 0.4, 0.6, 0.8, 1.0}
]
\addplot [thick, redisch, dotted, mark=*, mark size=1.5, mark options={solid}]
table [row sep=\\]{%
1 1 \\
2 0.77786446 \\
3 0.63531077 \\
4 0.54311067 \\
5 0.45677564\\
};
\end{axis}

\end{tikzpicture}
  }
  \caption{\rebuttal{Average response of all contact-frames in a recording session with respect to the average response of all contact-frames of session 1.}}
  \label{fig:sensor_degradation}
\end{figure}
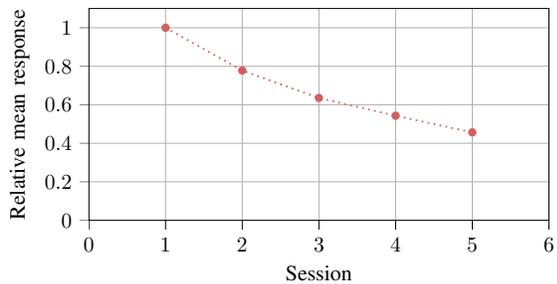



\begin{figure}[t]
\centering
  \resizebox{0.85\columnwidth}{!}{%
    \centering

\begin{tikzpicture}
\begin{axis}[
height=5cm,
width=9cm,
legend cell align={left},
legend style={at={(0.97,0.03)}, anchor=south east, draw=white!80.0!black},
tick align=outside,
tick pos=left,
x grid style={white!69.01960784313725!black},
xlabel={Session},
xmajorgrids,
xmin=0, xmax=5,
xtick style={color=black},
xtick = {0, 1, ..., 5},
y grid style={white!69.01960784313725!black},
ylabel={Relative mean response},
ymajorgrids,
ymin=0.0, ymax=1.1,
ytick = {0.0, 0.2, 0.4, 0.6, 0.8, 1.0}
]
\addplot [thick, redisch, dotted, mark=*, mark size=1.5, mark options={solid}]
table [row sep=\\]{%
1 1\\
2 0.87157613 \\
3 0.7810318 \\
4 0.7244892 \\
};
\addlegendentry{Square}
\addplot [thick, greyisch, dotted, mark=*, mark size=1.5, mark options={solid}]
table[row sep=\\]{%
1 1 \\
2 0.8167371 \\
3 0.6982073 \\
4 0.5872175 \\
};
\addlegendentry{Glove}
\end{axis}

\end{tikzpicture}
  }
  \caption{Response degradation of the sensor glove and a square control sensor.}
  \label{fig:degradation_comparison}
\end{figure}
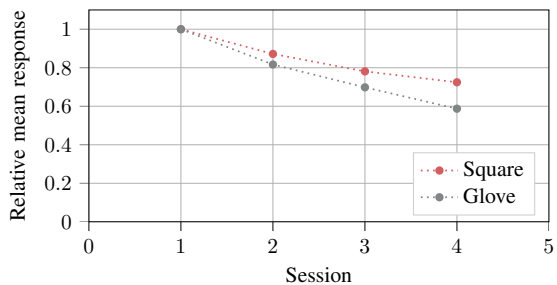

We did further evaluations using the fabricated control sensor explained in Sec.~\ref{subsec:methods:sensoreval}.

\subsubsection{Sensor degradation}
A possible cause for the large inter-session variability, apart from the general difficulty of repeatability, is that the response of the tactile data is in fact changing.
Indeed, analyzing the response of the sensor indicates that it is weaker for every successive recording session.
This behavior is observed when plotting the average across all contact-frames of each recording session and object class.
An example of such a plot is shown in Fig.~\ref{fig:comparison_class13}, clearly displaying a weaker response for later sessions.
\rebuttal{A more quantitative aspect is displayed in Fig.~\ref{fig:sensor_degradation}, where we plot the average response of all available contact-frames in each session with respect to the average response of all contact-frames of session 1.}
The weakening of the sensor response is monotonous and permanent and likely plays a big role in the inter-session variability presented in Fig.~\ref{fig:intersession}.
\rebuttal{We additionally observe that the highest inter-session accuracy is obtained when using the session 3 as validation set, as discussed in Sec.~\ref{sec:neuralnet}. Compared to the other sessions, the data from session 3 present intermediate response. This means that the model is trained using data with higher and lower values and validated on intermediate values. This has likely helped the model to generalize better on the session 3.}
}

\new{
\rebuttal{Finally, we compare the average response of the sensor glove and the control sensor to see if the degradation is caused by the extensive usage or if it is an intrinsic property of the materials over time.} 
Fig.~\ref{fig:degradation_comparison} shows the relative mean response compared to session 1 of the control sensor and the sensor glove. We can see that the degradation is similar for both sensor types. The only difference is that the response weakens slightly faster if the sensor is heavily used as in the case of the glove.
The decreasing response of the square sensor suggests that the degradation is not only caused by extensive use, but also simply by the combination of materials.

\subsubsection{\rebuttal{Slip evaluations}}

Thanks to the high frame rate of the analog front-end and the high spatial resolution of the sensor, a fast moving pressure spike is straightforward to capture.
An example frame sequence of such a case can be observed in Fig.~\ref{fig:screwdriver_slip}, where a screwdriver was pulled across the sensor surface.
It is much more difficult if the pressure is evenly distributed along the direction of movement.
If, additionally to being smooth and symmetrical, the moving object is longer than the sensor is wide it becomes impossible to recognize any movement by just inspecting the consecutive pressure frames.
This example is depicted in Fig.~\ref{fig:coke_slip}, where the `slipping' object is a full coke can.

\begin{figure}[b]
\captionsetup[subfigure]{labelformat=empty}
\centering
  \subfloat[]{\includegraphics[width=0.2\linewidth]{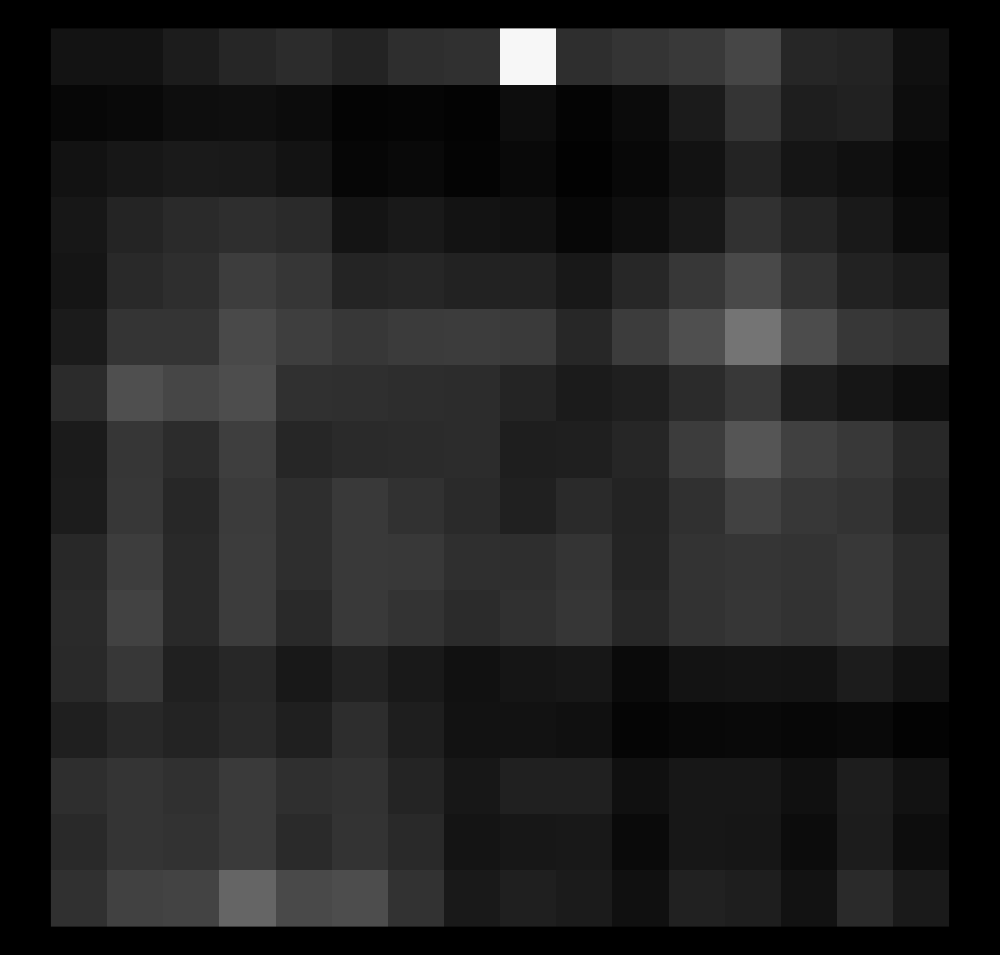}}\hfill
  \subfloat[]{\includegraphics[width=0.2\linewidth]{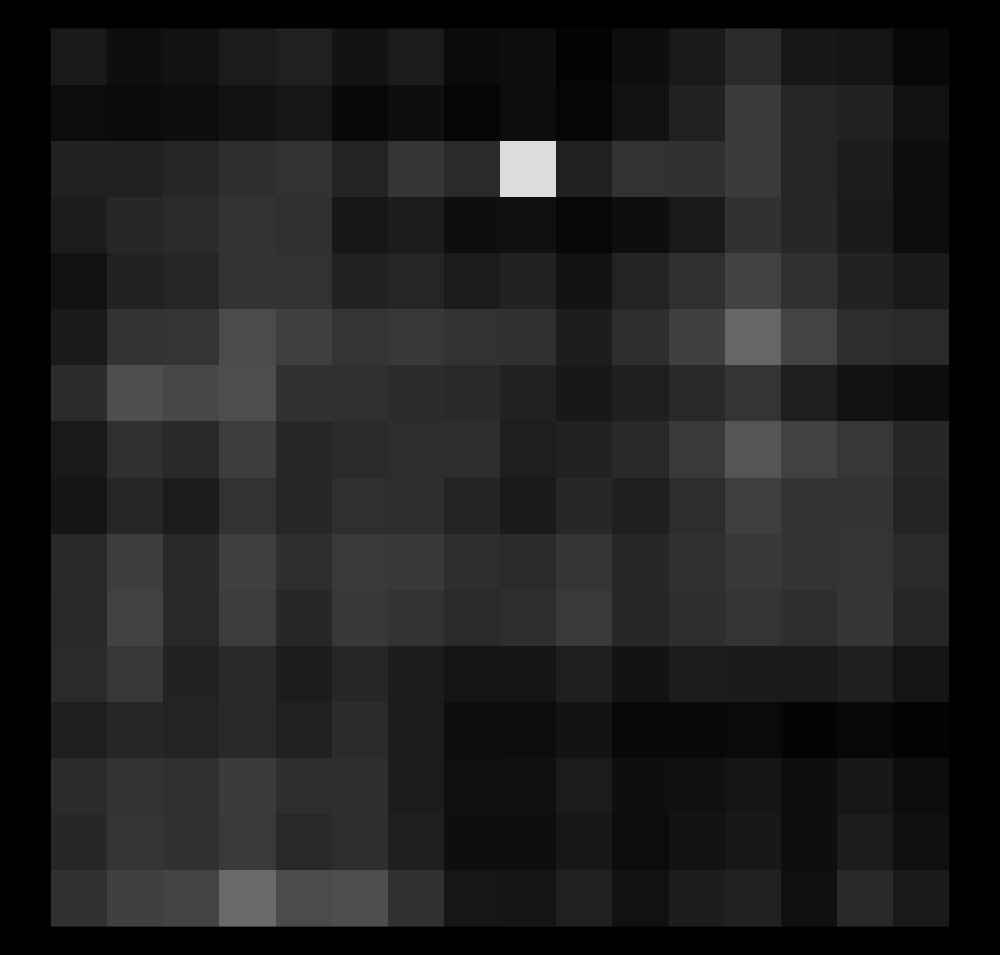}}\hfill
  \subfloat[]{\includegraphics[width=0.2\linewidth]{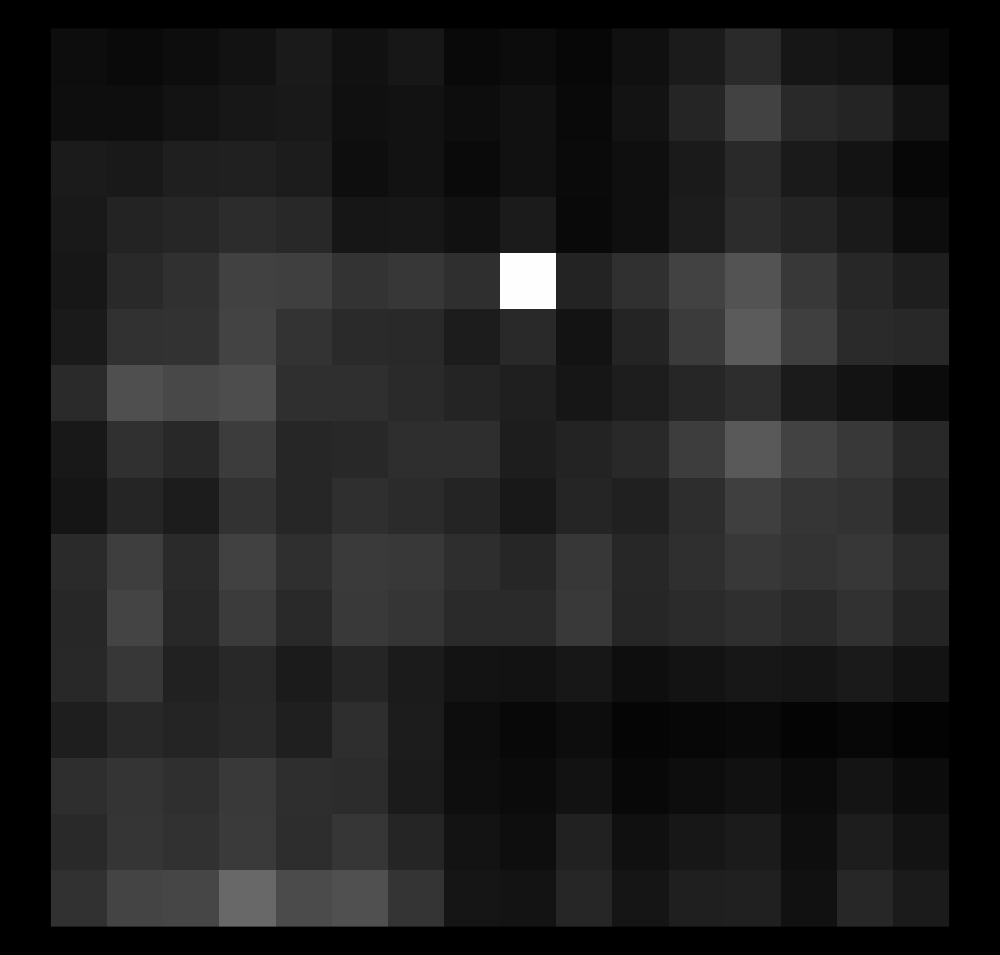}}\hfill
  \subfloat[]{\includegraphics[width=0.2\linewidth]{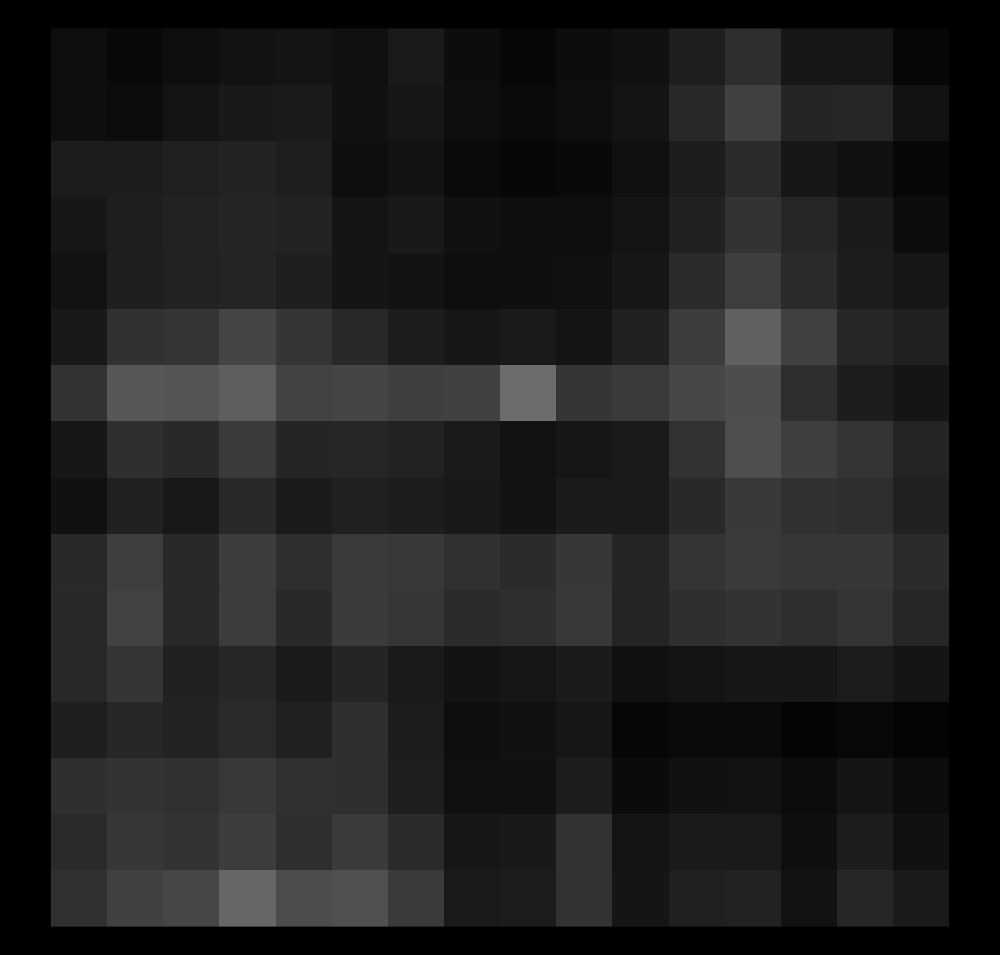}}\hfill
  \subfloat[]{\includegraphics[width=0.2\linewidth]{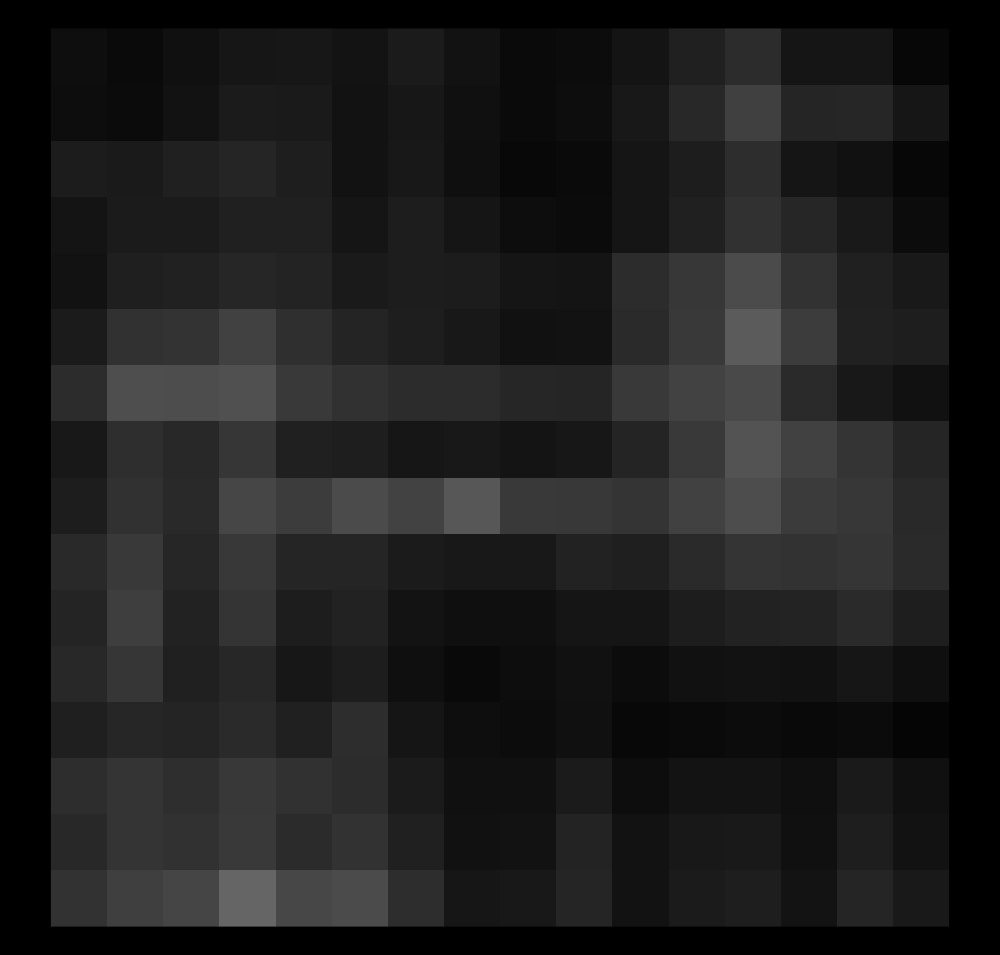}}
  \vspace{-0.5cm}
  \caption{Section of a frame sequence where a screwdriver was pulled from top to bottom across the sensor surface.}
  \label{fig:screwdriver_slip}
\end{figure}

\begin{figure}[b]
\captionsetup[subfigure]{labelformat=empty}
  \centering
  \subfloat[]{\includegraphics[width=0.2\linewidth]{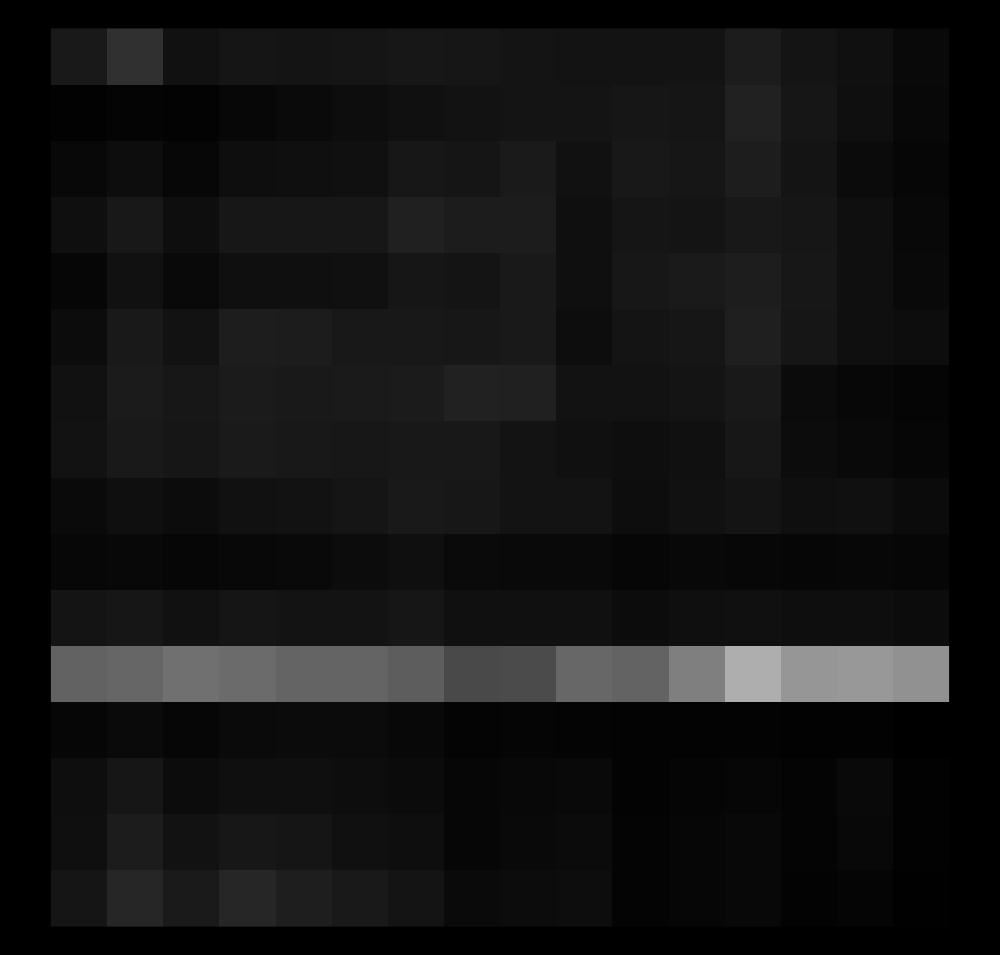}}\hfill
  \subfloat[]{\includegraphics[width=0.2\linewidth]{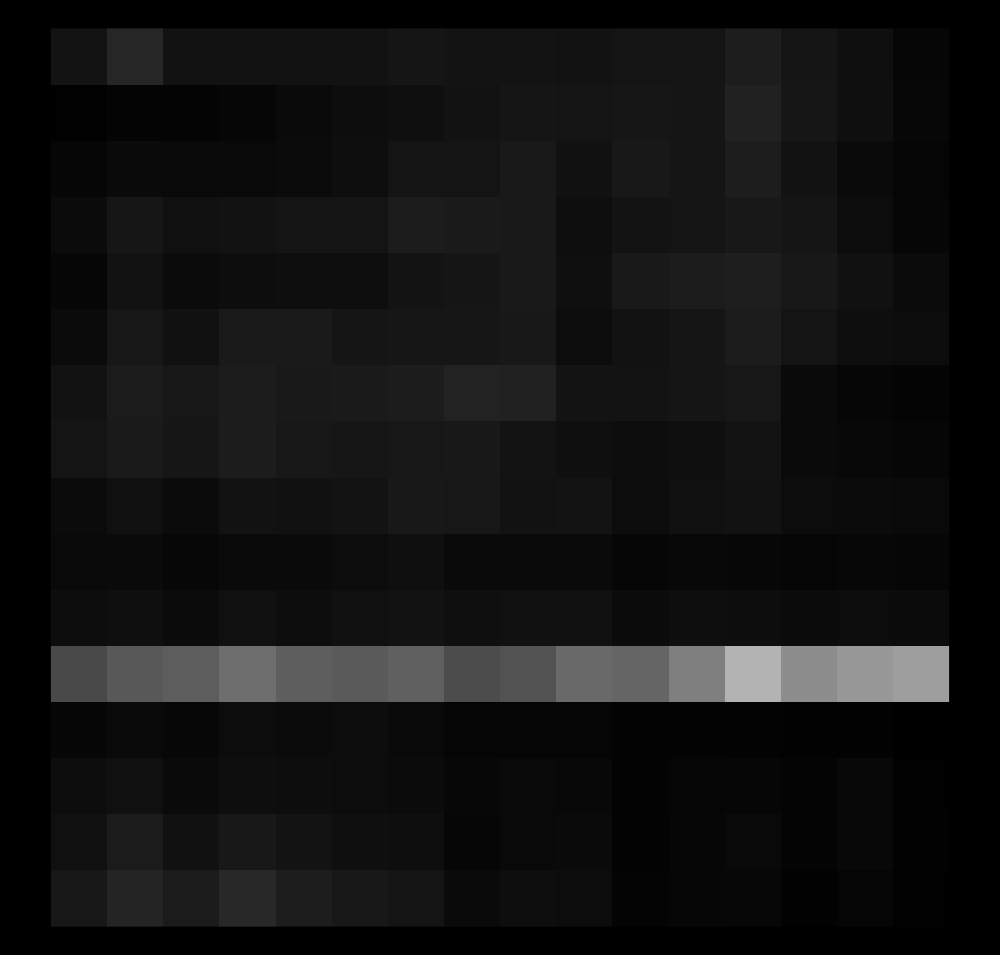}}\hfill
  \subfloat[]{\includegraphics[width=0.2\linewidth]{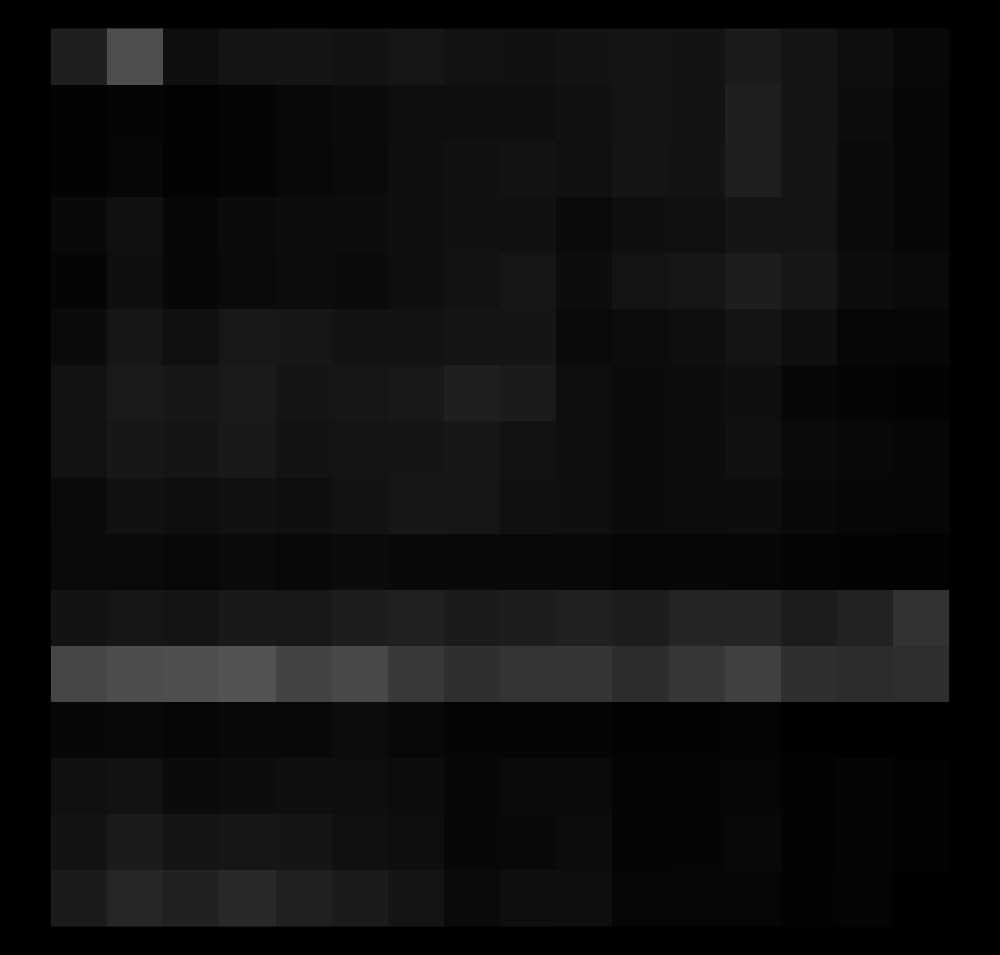}}\hfill
  \subfloat[]{\includegraphics[width=0.2\linewidth]{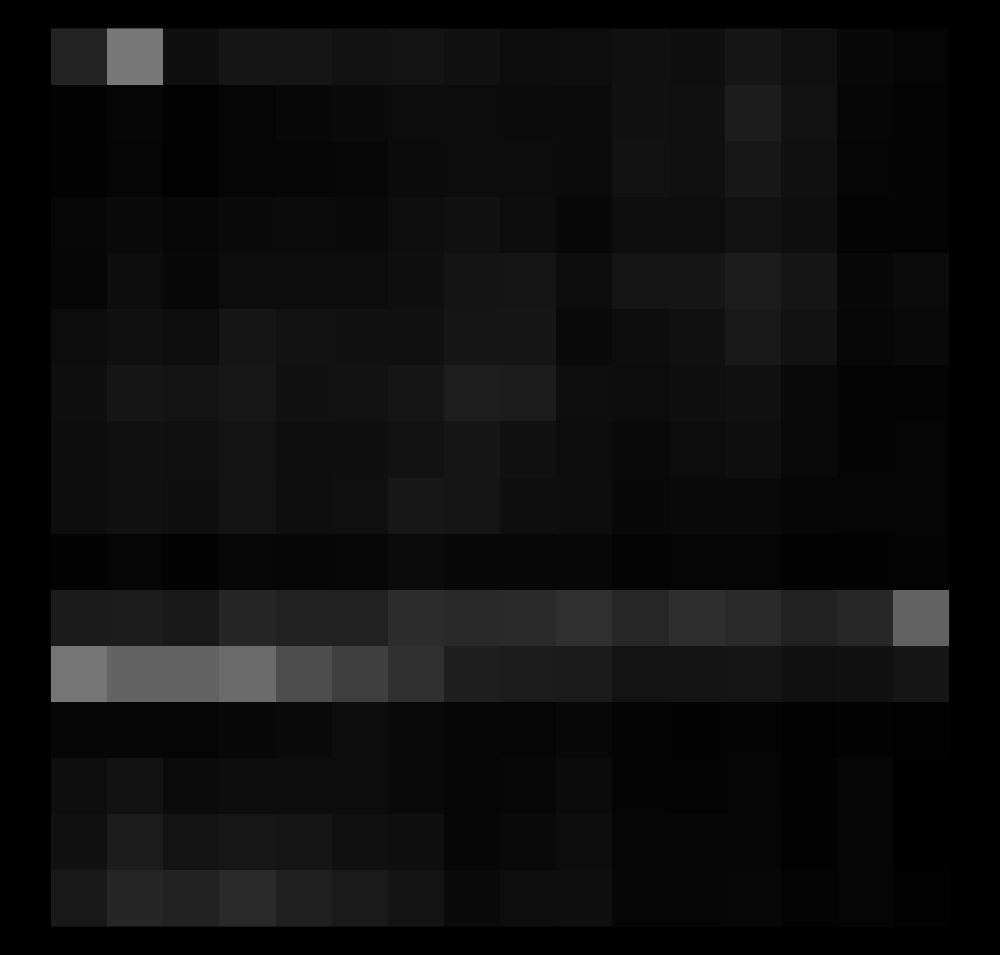}}\hfill
  \subfloat[]{\includegraphics[width=0.2\linewidth]{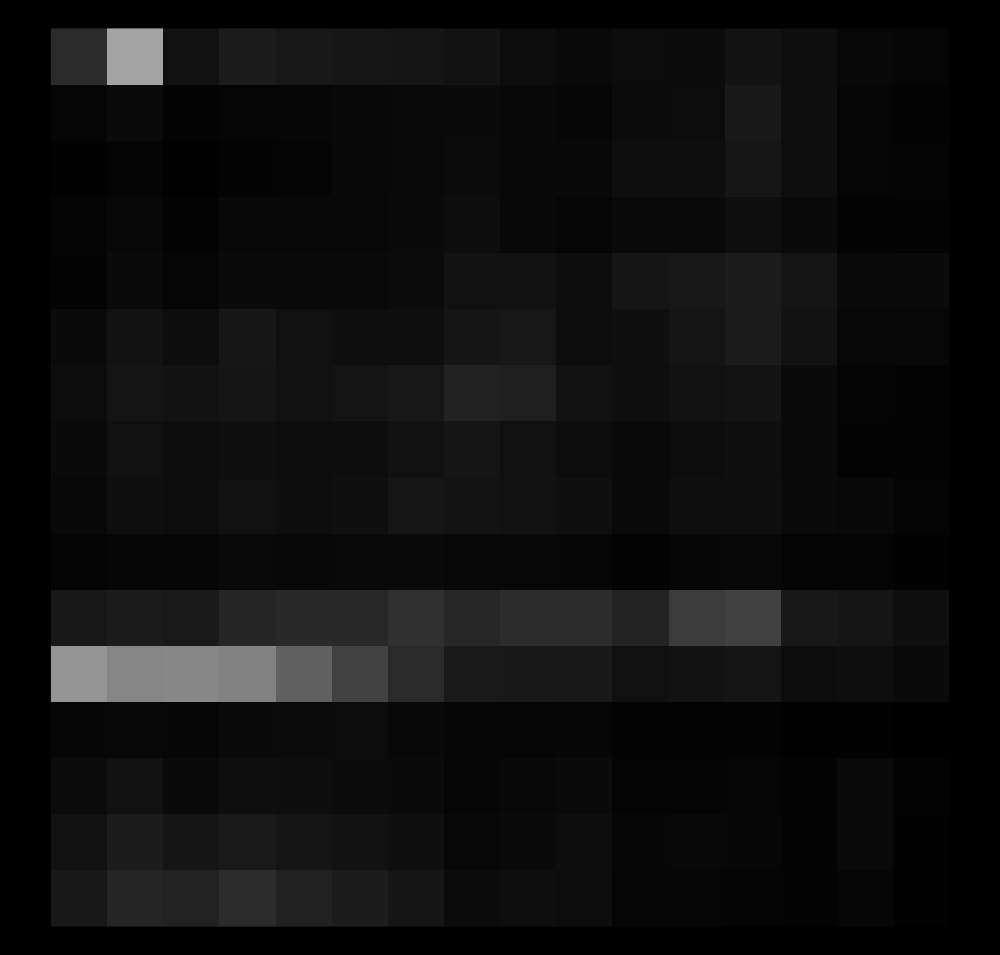}}
  \vspace{-0.5cm}
  \caption{Section of a frame sequence where a full coke can was pulled from left to right across the sensor surface.}
  \label{fig:coke_slip}
\end{figure}

\subsection{Discussions}\label{subsec:discussions}

With this work, we demonstrate a low-latency smart hand equipped with scalable tactile sensor and embedded machine learning able to classify different daily objects in real-time. 
\rebuttal{Possible applications include the real-time control of robotic or prosthetic hands or arms, where the embedded processing of the sensor data and its classification output can be immediately used to program subsequent movements. 
We achieve up to 98.86\% top-1 and 99.83\% top-3 accuracies when splitting the data from all sessions randomly. When a leave-one-session-out validation is performed, the average top-1 and and top-3 inter-session accuracy drops to 49.63\% and 77.84\%, respectively, with the highest validation results being 49.63\% top-1 and 77.84\% top-3 when the data from the intermediate session is used as validation set.}

\rebuttal{We perform additional sensor evaluations and show that the degradation of the tactile sensor does not originate only from repetitive usage, but is also an intrinsic property of the combination of materials (electrodes and adhesive).}
Another appropriate route of exploration is the continuous use and examination of the sensor glove.
The degradation in Fig.~\ref{fig:sensor_degradation} slows down and shows a convergence to a stable behavior with further use.
If the sensor degradation effectively stops, the analog front-end can easily be tuned to the final stable response and it is reasonable to expect large accuracy gains.

\rebuttal{
Finally, the dataset used in this work is collected by placing the objects on a table and handling them mostly from the top. This first step of recognizing an object is important for a robotic arm to decide how to better handle the object in the following steps, for example, to lift it. In this subsequent scenario, slip detection becomes relevant. Our experiments on slip evaluations suggest that it is possible to observe the slipping motion of an object using this sensor. In future work, a new dataset can be collected, for example, while lifting or moving the objects from one location to another, to evaluate slip detection algorithms. Likewise, the utility of the IMU data can be further assessed in these scenarios where the handling of the objects is more variable. Finally, the proposed system can be embedded on robotic hands for assessing its performance on-field, and the combination with other modalities such as electromyography~\cite{Meattini2018sEMG} can be explored.
}
}

\section{Conclusion}
\label{ch:conclusion}

\rebuttal{This paper presents SmartHand, a smart embedded system that is a step towards equipping robotic and prosthetic devices with a sense of touch. 
Starting from replicating a state-of-art tactile sensor with high spatial resolution, SmartHand focuses on maximizing the temporal resolution by reducing the latency in the system response. This is made possible thanks to an improved analog front-end and the real-time execution of a compact deep learning model embedded close to the sensor node.}
A working prototype of SmartHand is designed and implemented to carry out solid experimental evaluations that have shown an overall power consumption of 505\,mW and 185\,\textmu W respectively in active and standby mode, and a response latency of 100\,ms achieving an inter-session accuracy of 77.84\% (top-3) in classifying 17 classes, i.e., 16 objects and the empty hand. \rebuttal{Further sensor evaluations demonstrate the changing sensor response with repetitive usage and with sliding objects, providing insights for future developments.}

\bibliographystyle{IEEEtran}
\bibliography{bib}

\begin{IEEEbiography}[{\includegraphics[width=1in,height=1.25in,clip,keepaspectratio]{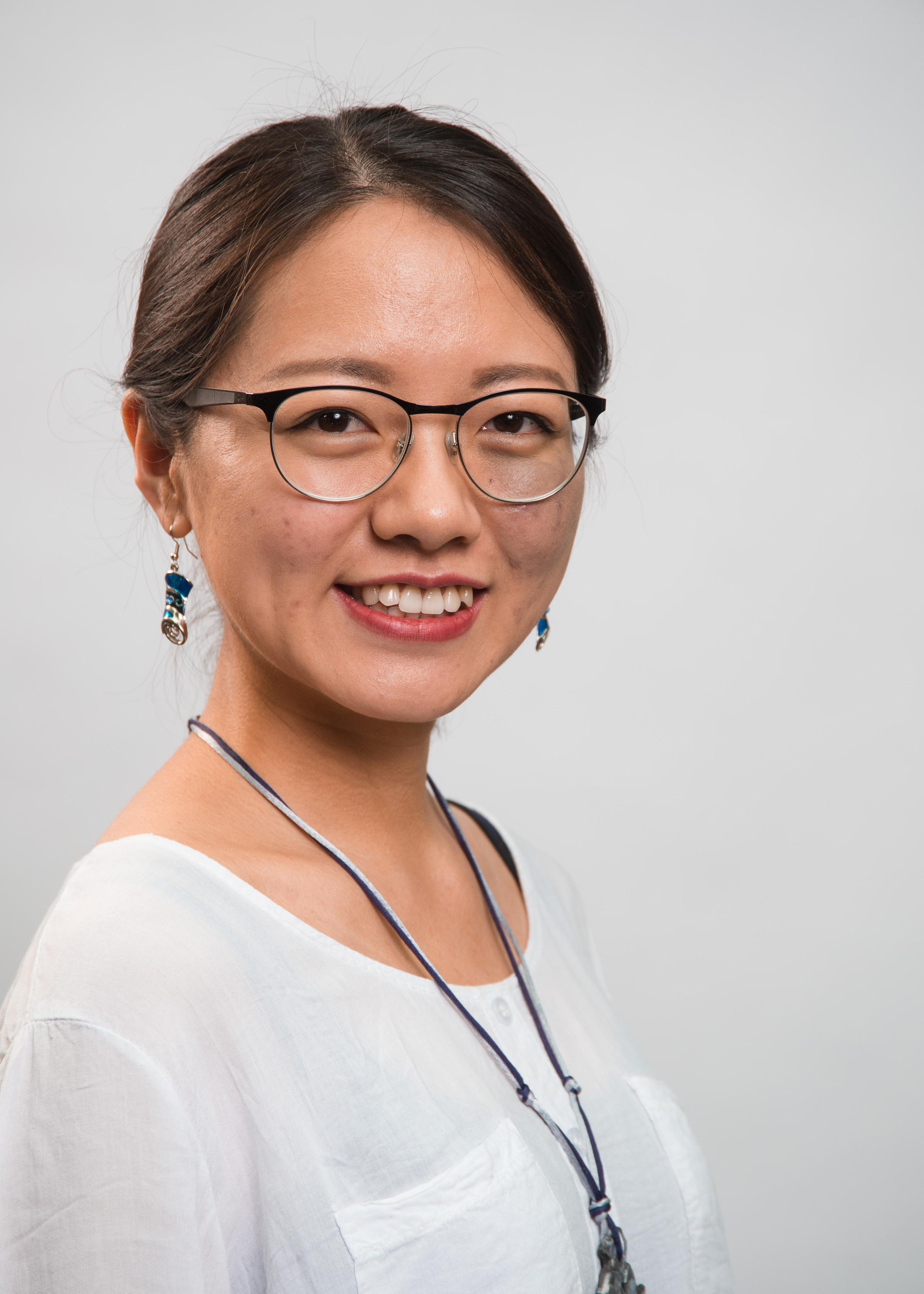}}]{Xiaying Wang}
received her B.Sc. and M.Sc. degrees in biomedical engineering from Politecnico di Milano, Italy and ETH Zürich, Switzerland in 2016 and 2018, respectively. She is currently pursuing a Ph.D. degree at the Integrated Systems Laboratory at ETH Zürich. Her research interests include biosignal processing, low power embedded systems, energy-efficient smart sensors, brain—machine interfaces, and machine learning on microcontrollers. She received the excellent paper award at the IEEE Healthcom conference in 2018 and she won the Ph.D. Fellowship funded by Swiss Data Science Center in 2019. 
\end{IEEEbiography}

\begin{IEEEbiography}[{\includegraphics[width=1in,height=1.25in,clip,keepaspectratio]{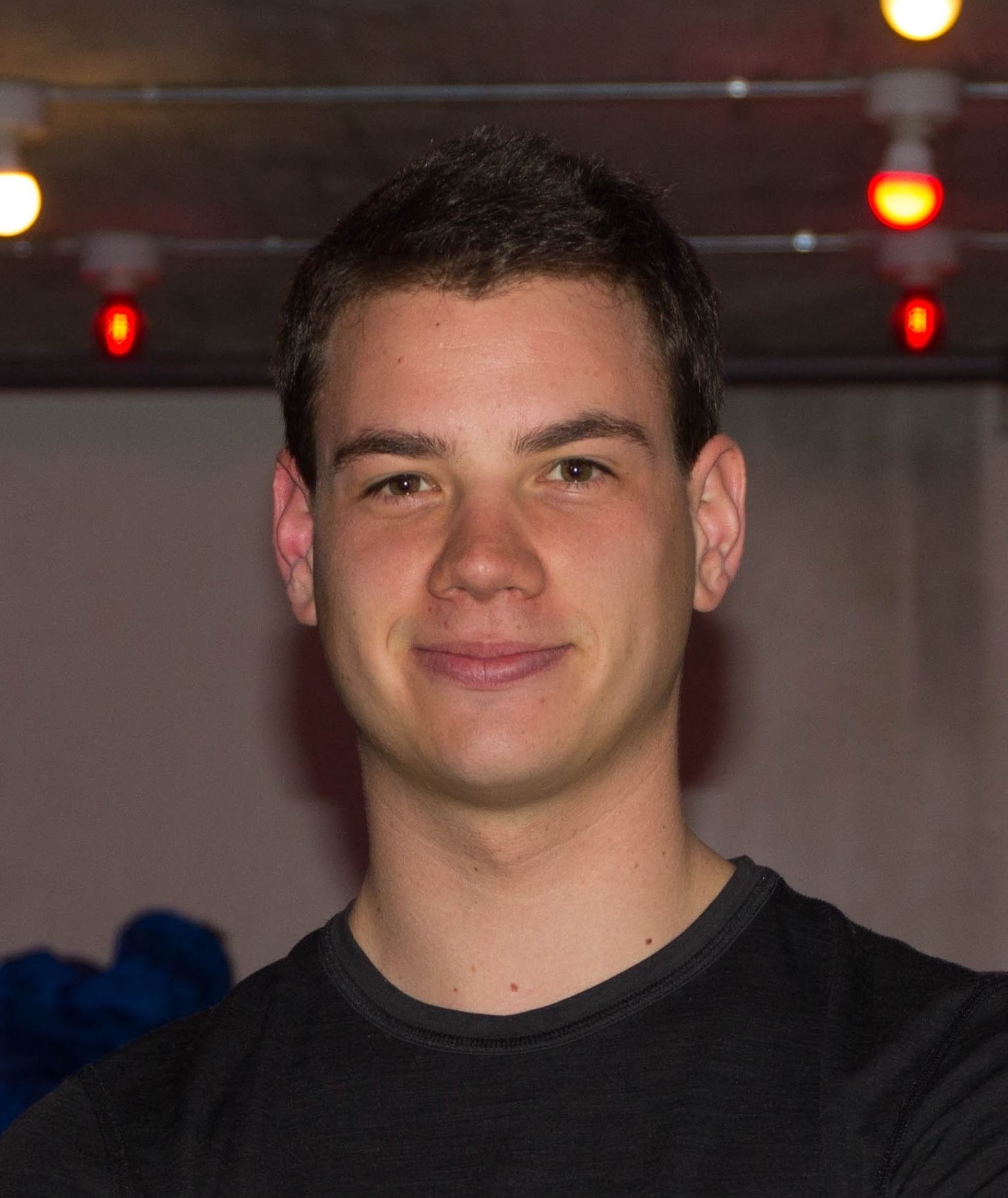}}]{Fabian Geiger}
received his B.Sc. degree in Electrical Engineering and his M.Sc. degree in Biomedical Engineering from ETH Zürich, Switzerland in 2018 and 2020, respectively.
He is currently working as a design engineer in the fields hardware and firmware. His interests include edge computing, smart sensor nodes, machine learning on microcontrollers and tinkering with new technologies.
\end{IEEEbiography}

\begin{IEEEbiography}[{\includegraphics[width=1in,height=1.25in,clip,keepaspectratio]{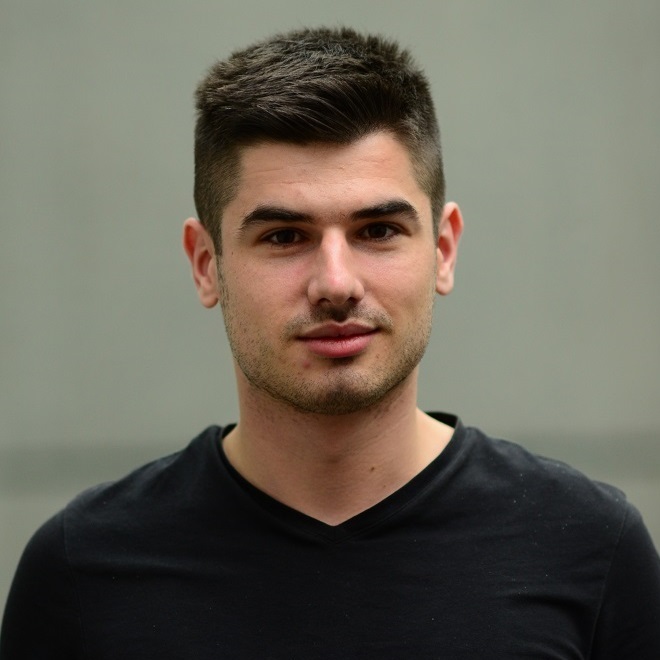}}]{Vlad Niculescu}
received the Master's degree in Robotics, Systems, and Control from the ETH Zürich, in 2019. He is currently pursuing the Ph.D. in Electrical Engineering within the Integrated Systems Laboratory in ETH Zürich. During the Bachelor and Master period, he competed in more than ten international student competitions, and he was the electrical lead of the student project Swissloop, which won second place and the innovation award in the SpaceX Hyperloop Pod Competition 2019. His research is now focused on developing localization and autonomous navigation algorithms that target ultra-low-power platforms which can operate onboard nano-drones.
\end{IEEEbiography}

\begin{IEEEbiography}[{\includegraphics[width=1in,height=1.25in,clip,keepaspectratio]{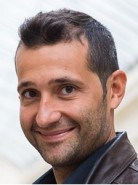}}]{Michele Magno}
is currently a Senior Researcher and Lecturer at ETH Zurich, Switzerland, at the Department of Information Technology and Electrical Engineering (D-ITET). Since 2020, he is leading the D-ITET center for project-based learning. He received his master and Ph.D. degrees in electronic engineering from the University of Bologna, Italy, in 2004 and 2010, respectively. He is working in ETH since 2013 and has become a visiting lecturer or professor in several universities, namely the University of Nice Sophia, France, Enssat Lannion, France, University of Bologna and Mid University Sweden. His current research interests include smart sensing, low power machine learning, wireless sensor networks, wearable devices, energy harvesting, low power management techniques, and extension of the lifetime of batteries-operating devices. He has authored more than 200 papers in international journals and conferences. Some of his publications were awarded as best papers awards in IEEE conferences.
\end{IEEEbiography}

\begin{IEEEbiography}[{\includegraphics[width=1in,height=1.25in,clip,keepaspectratio]{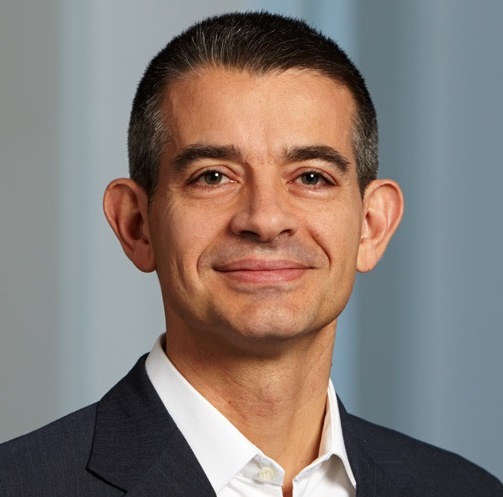}}]{Luca Benini}
is the Chair of Digital Circuits and Systems at ETH Zürich and a Full Professor at the University of Bologna. He has received a PhD from Stanford University and has served as Chief Architect for the Platform2012 in STMicroelectronics, Grenoble. Dr. Benini’s research interests are in energy-efficient system and multi-core SoC design. He is also active in the area of energy-efficient smart sensors and sensor networks. He has published more than 1000 papers in peer-reviewed international journals and conferences, four books and several book chapters. He is a Fellow of the ACM and of the IEEE and a member of the Academia Europaea.
\end{IEEEbiography}

\end{document}